%% file: main.tex
\documentclass[10pt,twocolumn,letterpaper]{article}
\PassOptionsToPackage{table}{xcolor}

\usepackage[pagenumbers]{cvpr} % To force page numbers, e.g. for an arXiv version

\usepackage{booktabs}
\usepackage{makecell}
\usepackage{tabularx}
\usepackage{subcaption}
\usepackage{algorithm}
\usepackage{caption}
\usepackage{algpseudocode}
\usepackage{bibunits}
\definecolor{cvprblue}{rgb}{0.21,0.49,0.74}
\usepackage[pagebackref,breaklinks,colorlinks,allcolors=cvprblue]{hyperref}
\definecolor{oursgray}{gray}{0.92}
\def\confName{CVPR}
\def\confYear{2027}
\usepackage{microtype}
\title{Attribute-Aware Crowd Generation\vspace{-4mm}}
\author{JunGyu Lee$^{1,*}$ \quad Chaemin Hwang$^{1,*}$ \quad Jisu Shin$^{2}$ \quad Seunghyun Shin$^{2}$ \quad Hae-Gon Jeon$^{1,\dagger}$\\
$^{1}$Yonsei University \quad $^{2}$GIST \quad {\normalsize $^{*}$Equal contribution \quad $^{\dagger}$Corresponding author}\\[-0.2em]
{\small\href{https://jungyu0413.github.io/ACG/}{\texttt{https://jungyu0413.github.io/ACG/}}}}
\begin{document}

\twocolumn[{%
\renewcommand\twocolumn[1][]{#1}%
\maketitle

\vspace{-8mm}          % 기존 -6mm, 링크와 그림 사이 간격
{\centering
\includegraphics[width=\textwidth]{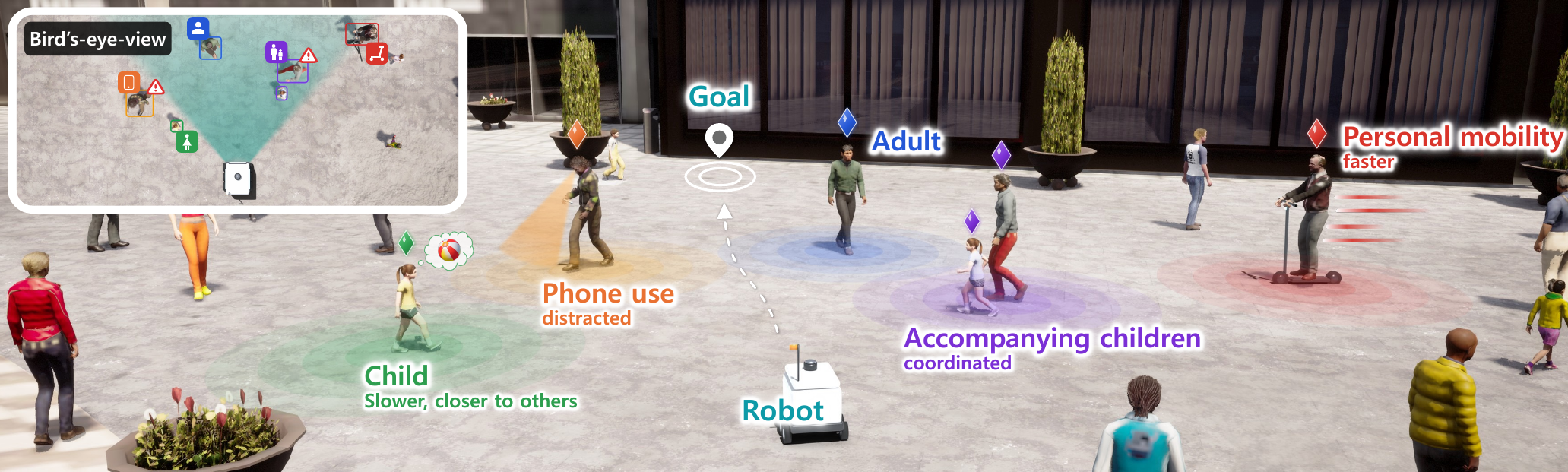}

\vspace{-0.5em}

\captionsetup{hypcap=false}
\captionof{figure}{
\textbf{Attribute-aware crowd generation.}
People with different attributes move and interact in different ways. Our model is the first attempt to simulate pedestrians’ motions and interactions according to their attributes such as age, accompanying children, and phone use.}
\label{fig:teaser}
\par}

\vspace{0.3em}
}]

\input{sec/0_abstract}
\input{sec/1_intro}
\input{sec/2_related_work}
\input{sec/3_Dataset}
\input{sec/4_method}
\input{sec/5_experiments}
\input{sec/6_conclusion}

\section*{LLM Usage Statement}
We used ChatGPT and Claude to edit the grammar of the paper and to assist with writing code for baseline methods. We did not use them to develop the research idea, conduct the experiments and comparisons, or analyze the results. We reviewed all text and code made with these tools and took full responsibility for the final content.

{
    \small
    \bibliographystyle{ieeenat_fullname}
    \bibliography{main}
}
\begin{bibunit}[ieeenat_fullname]
    \backrefsetup{disable}
    \input{sec/X_suppl}

    \renewcommand{\refname}{Supplementary References}
    {\small\putbib[main]}
\end{bibunit}

\end{document}

%% file: sec/0_abstract.tex
\begin{abstract}
\par
\vspace{-3mm}
Crowds differ from place to place, and behavior differs from person to person. For example, children often walk with their parents around schools, while phone users tend to pay less attention to their surroundings. Capturing this diversity is essential for realistic crowd simulation. 
However, since datasets that provide both scene maps and pedestrian attribute annotations remain limited, existing crowd generation models do not explicitly learn attribute-dependent patterns. In this paper, we introduce an \textbf{attribute-aware crowd generation pipeline}. To do this, we first construct a crowd trajectory dataset with world-coordinate trajectories, scene semantic maps, and annotations for pedestrian attributes selected based on social studies.
Building on this dataset, we propose a framework that models how attribute distributions vary across scenes and how these attributes form agents' motion patterns.
Here, we incorporate two attention mechanisms: a scene cross-attention for predicting attributes from agent origins and scene maps, and an attribute-aware social attention for capturing the relative influence of each neighbor based on both agents' attributes and their relative positions. 
Extensive experiments demonstrate that our model outperforms baselines on public datasets as well as our attribute-annotated dataset. As an application, robot navigation policies trained on our generated crowds achieve lower collision rates while maintaining comparable success rates. To the best of our knowledge, our pipeline is the first attempt to enable attribute-aware crowd generation.

\end{abstract}

%% file: sec/1_intro.tex
\section{Introduction}
\label{sec:intro}

Crowd simulation supports the design and evaluation of autonomous driving systems, mobile robots, and urban environments. For example, a mobile robot navigating a park must move safely among adults accompanying young children, cyclists, and pedestrians using phones. Evaluating its safety requires considering both the composition of pedestrians in the park and their movement patterns and interactions. However, since reproducing real-world conditions is very difficult, crowd simulation is needed to provide diverse scenarios for training and evaluation under controlled conditions.

To capture this diversity, we focus on two key aspects. First, pedestrian composition varies across environments. Older adults are more common on park walking loops than in other park areas~\cite{cohen2016first}. Near primary schools, crowds include children walking with parents or guardians~\cite{zacharias2017local}. Pedestrians using phones are also more common near a university campus than in entertainment districts~\cite{schwebel2022distracted}. Second, pedestrian motions and interactions vary with individual attributes. People of different ages tend to walk at different speeds and behave differently when avoiding collisions~\cite{morrongiello2024developmental,willis2004human,rapos2019minimum}. Adults walking with children adopt different group formations from adult-only groups~\cite{zanlungo2017intrinsic}. Phone use can also interfere with anticipating others' movements and coordinating avoidance~\cite{murakami2021mutual,murakami2022spontaneous}. These studies motivate an attribute-aware crowd simulation framework that models both aspects.

However, existing crowd generation methods~\cite{reynolds,helbing,rvo,spdiff,trace,crowdes} have difficulty in modeling these two aspects. Although CrowdES~\cite{crowdes} generates agent types, walking speeds, and appearance times, it does not explicitly learn how pedestrian attributes are distributed across scenes or influence behavior. Furthermore, learning these behavioral patterns requires trajectories annotated with individual attributes and scene context. Unfortunately, existing public datasets provide annotations only for age-related attributes~\cite{panda,gigatraj,jahangard2024jrdb}. Therefore, we need a dataset that captures scene context, a broad range of pedestrian attributes, and walking behavior, along with a model that learns from it to reproduce diversity in both composition and behavior.

In this work, we introduce an attribute-aware crowd generation pipeline. We construct \textbf{CAT} (\textbf{C}rowd \textbf{A}ttribute and \textbf{T}rajectory dataset) from high-resolution PANDA videos~\cite{panda}. Based on studies of pedestrian behaviors and social interactions~\cite{zanlungo2017intrinsic, morrongiello2024developmental,murakami2022spontaneous}, we select four attribute groups identifiable in videos: age, mobility mode, accompanying children, and phone use. We manually annotate these attributes and scene semantic maps, and convert image-coordinate trajectories into world coordinates using scene-specific homographies.

Building on CAT, we propose \textbf{ACG} (\textbf{A}ttribute-aware \textbf{C}rowd \textbf{G}eneration) for continuous crowd generation using scene information derived from a single real-world image, following CrowdES~\cite{crowdes}. Our model consists of an emission module that sets agents' initial states, including their starting positions and goals, and a motion module that generates crowd motions. We incorporate pedestrian attributes into both modules through two attention mechanisms. First, during emission, a scene cross-attention predicts attribute distributions from each sampled origin and the scene map. The sampled attributes guide initial goal selection. Second, during motion generation, an attribute-aware social attention computes each neighbor's influence from both agents' attributes and their relative positions, allowing attention weights to vary with attributes even at identical relative positions. The motion module infers each agent's next displacement autoregressively. Both modules share a scene encoder for scene semantics and layout, and are trained jointly.

% Experiments on CAT show that ACG captures attribute proportions and attribute-dependent walking patterns. Ablation studies further show that the two attention modules contribute to capturing these attribute-related characteristics. Our model also reproduces realistic crowd statistics and trajectory patterns on public benchmarks such as ETH~\cite{eth}/UCY~\cite{ucy}, SDD~\cite{sdd}, and GCS~\cite{GCS,gcs2}. Furthermore, robot navigation policies trained on ACG-generated crowds achieve lower collision rates while maintaining comparable success rates. Thanks to the benefits of our attribute-aware crowd simulation, robot navigation policies also perform well in unseen scenes and with crowds generated by off-the-shelf models.

To the best of our knowledge, ACG is the first crowd generation pipeline to model scene-dependent pedestrian composition and attribute-dependent behavior. Experiments demonstrate that ACG reproduces real-world crowd statistics and trajectory patterns on public benchmarks such as ETH~\cite{eth}/UCY~\cite{ucy}, SDD~\cite{sdd}, and GCS~\cite{GCS,gcs2}, as well as CAT. Quantitative and qualitative analyses on CAT further show that our model captures attribute proportions and attribute-dependent walking patterns. Ablation studies confirm the contribution of the two attention mechanisms to capturing these attribute-related patterns. Thanks to our attribute-aware crowd simulation, robot navigation policies trained on ACG-generated crowds achieve lower collision rates while maintaining comparable success rates and also perform well in unseen scenes and with crowds generated by off-the-shelf models.

%  We confirm that CAT has crowd statistics and trajectory patterns similar to real-world ones. Quantitative and qualitative analyses on CAT show that ACG captures attribute proportions and attribute-dependent walking patterns. Ablation studies demonstrate the effectiveness of two attention mechanisms. In addition, our model reproduces real-world crowd statistics and trajectory patterns well on public benchmarks such as ETH~\cite{eth}/UCY~\cite{ucy}, SDD~\cite{sdd}, and GCS~\cite{GCS,gcs2}, as well as CAT.  Furthermore, robot navigation policies trained on ACG-generated crowds achieve lower collision rates while maintaining comparable success rates. Thanks to the benefits of our attribute-aware crowd simulation, robot navigation policies also perform well in unseen scenes and with crowds generated by off-the-shelf models. To the best of our knowledge, our pipeline is the first to model both scene-dependent pedestrian composition and attribute-dependent behavior.

%% file: sec/2_related_work.tex
\section{Related Work}
\label{sec:related_work}

\begin{figure*}[t]
    \centering
    \includegraphics[width=\textwidth]{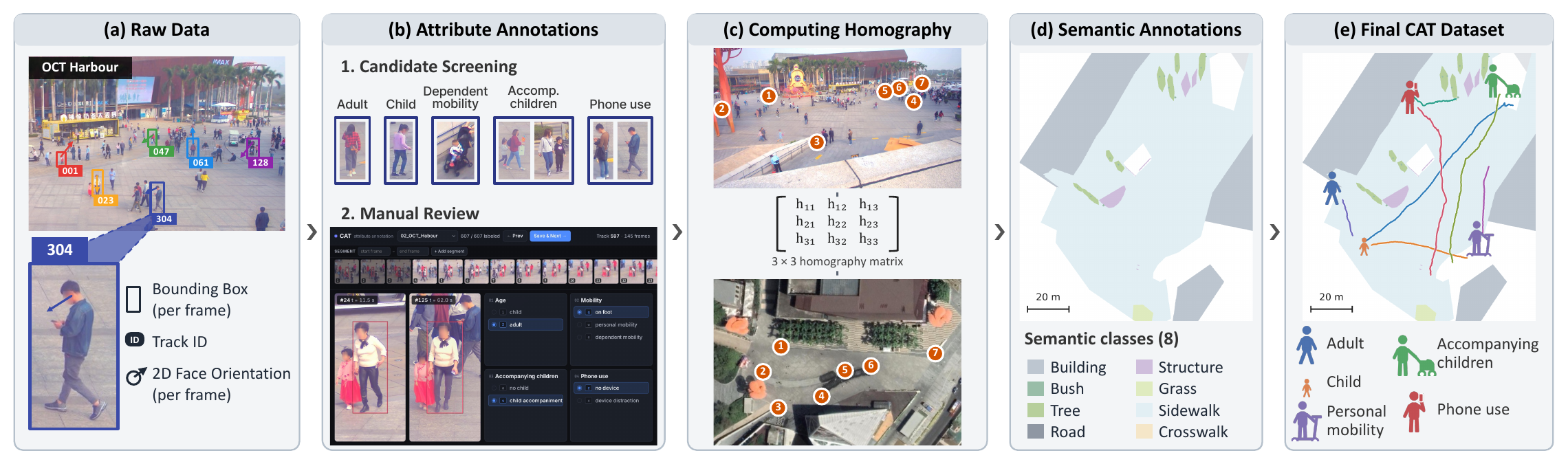}
    \vspace{-6mm}
    \caption{
    \textbf{CAT construction pipeline.} (a) Raw data provide pedestrian tracks and visual cues. (b) We screen candidate pedestrians and refine labels using a web-based annotation tool. (c) We reconstruct world-coordinate trajectories using homographies. (d) We construct scene semantic maps using Grounded-SAM and manual labeling. (e) CAT combines pedestrian attributes, trajectories, and semantic maps.
    }
    \vspace{-3mm}
    \label{fig:cat_pipeline}
\end{figure*}

\subsection{Crowd Trajectory Datasets}

Real-world pedestrian trajectory datasets support training and evaluation of models capturing pedestrian motions and interactions across diverse environments~\cite{eth,ucy,sdd,GCS,edinburgh,ind,brvsvcic2013person,zanlungo2014potential,rudenko2020thor,pets2009,benfold2011stable}. Autonomous driving and robot navigation datasets provide pedestrian trajectories and annotations for perception and behavior analysis~\cite{kitti,nuscenes,waymo,waymo_motion,jrdb,jaad,pie,titan,bdd100k,l-cas}. PANDA~\cite{panda} annotates pedestrian posture for large-scale crowd analysis. GigaTraj~\cite{gigatraj} combines world-coordinate trajectories with semantic scene maps, while JRDB-Social~\cite{jahangard2024jrdb} includes individual attribute and social group annotations. Despite these contributions, publicly available datasets do not provide annotations for attributes relevant to behavior, such as age, mobility mode, accompanying children, and phone use.

\subsection{Multi-Agent Trajectory Prediction}

Multi-agent trajectory prediction forecasts motion from past trajectories, interactions, and scene context. Existing methods model social interactions using pooling~\cite{social_lstm,social_gan}, attention~\cite{sophie,stgat,bigat}, graph networks~\cite{social_stgcnn,sgcn}, and transformers~\cite{star,agentformer}. Scene-aware methods use environmental context~\cite{sophie,trajectronpp,ynet}, while goal-based methods condition on destinations~\cite{pecnet,goal_gan,SGNet,ynet}. Studies address multimodal motion~\cite{mid,memonet} and interactions at individual and group levels~\cite{grouptron,groupnet,gpgraph}. Recent work explores higher-order social interactions~\cite{pchgcn}, trajectory spectra~\cite{resonance}, flow-based prediction~\cite{moflow}, and language-based representations~\cite{lmtraj,w2w}. While these methods generate realistic motion, their dependence on past trajectories limits application to crowd generation.

\subsection{Crowd Simulation and Generation}

% Traditional crowd simulation methods use hand-crafted rules for flocking~\cite{reynolds}, social forces~\cite{helbing}, and reciprocal collision avoidance~\cite{rvo}, while continuum methods model collective crowd flow~\cite{continuum}. Example-based methods reuse recorded motion, while interactive approaches control crowds through navigation fields, proxy agents, and animation editing~\cite{lee2007group, ucy,ju2010morphable,patil2010directing,yeh2008composite,colas2022interaction,kim2014interactive}. Learning-based approaches model pedestrian navigation and behavior through reinforcement learning and behavioral controls~\cite{lee2018crowd,greilcrowds,panayiotou2022ccp,cedrl}. These methods simulate crowds in predefined environments rather than generating crowds directly from real-world scene images. Recent generative methods include social-force-guided diffusion in SPDiff~\cite{spdiff}, social-physics-informed diffusion~\cite{envsocial}, guided diffusion in TRACE~\cite{trace}, text-guided crowd generation~\cite{ji2024text}, and continuous agent emission and motion generation with population and agent controls in CrowdES~\cite{crowdes}. Our work focuses on where pedestrians with different attributes appear across scenes and how these attributes influence walking behavior.

Traditional crowd simulation methods use hand-crafted rules for flocking~\cite{reynolds}, social forces~\cite{helbing}, and reciprocal collision avoidance~\cite{rvo}, while continuum methods model collective crowd flow~\cite{continuum}. Example-based methods reuse recorded motion, while interactive approaches control crowds through navigation fields, proxy agents, and animation editing~\cite{lee2007group,ucy,ju2010morphable,patil2010directing,yeh2008composite,colas2022interaction,kim2014interactive}. Learning-based approaches model pedestrian navigation and behavior through reinforcement learning and behavioral controls~\cite{lee2018crowd,greilcrowds,panayiotou2022ccp,cedrl}. Recent generative methods include social-force-guided diffusion in SPDiff~\cite{spdiff}, social-physics-informed diffusion~\cite{envsocial}, guided diffusion in TRACE~\cite{trace}, and text-guided crowd generation~\cite{ji2024text}. CrowdES~\cite{crowdes} supports continuous crowd generation from real-world scene images, combining agent emission and motion generation with population and agent controls. Our work focuses on where pedestrians with different attributes appear across scenes and how these attributes influence walking behavior.

% Traditional crowd simulation methods use hand-crafted rules for flocking~\cite{reynolds}, social forces~\cite{helbing}, and reciprocal collision avoidance~\cite{rvo}, while continuum methods model collective crowd flow~\cite{continuum}. Example-based methods reuse recorded motion, while interactive approaches control crowds through navigation fields, proxy agents, and animation editing~\cite{lee2007group, ucy,ju2010morphable,patil2010directing,yeh2008composite,colas2022interaction,kim2014interactive}. Learning-based approaches model pedestrian navigation and behavior through reinforcement learning and behavioral controls~\cite{lee2018crowd,greilcrowds,panayiotou2022ccp,cedrl}. Recent generative methods include social-force-guided diffusion in SPDiff~\cite{spdiff}, social-physics-informed diffusion~\cite{envsocial}, guided diffusion in TRACE~\cite{trace}, text-guided crowd generation~\cite{ji2024text}, and continuous agent emission and motion generation with population and agent controls in CrowdES~\cite{crowdes}.

% Our work focuses on how attributes relate to pedestrian composition and walking behaviors. The scene cross-attention predicts attribute distributions conditioned on origins and scene context, while the attribute-aware social attention models interactions based on pedestrian attributes.

%% file: sec/3_Dataset.tex
\section{The CAT Dataset}
\label{sec:dataset}

We introduce CAT, a Crowd Attribute and Trajectory dataset with attribute annotations. CAT combines world-coordinate trajectories with scene semantic maps and pedestrian attribute annotations. These annotations cover four visually identifiable attribute groups that influence pedestrian behaviors~\cite{zanlungo2017intrinsic, morrongiello2024developmental,murakami2022spontaneous}: age, mobility mode, accompanying children, and phone use. By aligning these attributes with trajectories and scene context, our dataset supports the modeling and evaluation of scene-dependent pedestrian composition and attribute-dependent motion patterns. Fig.~\ref{fig:cat_pipeline} illustrates the dataset construction pipeline. Additional construction details are provided in Appendix A.
% ~\ref{app:cat_details}

\subsection{Data Collection}
\label{sec:data_collection}
CAT is constructed from six video sequences in PANDA~\cite{panda}, a gigapixel-level dataset for large-scale human-centric visual analysis. We obtain the videos, with faces de-identified, and their tracking annotations through the GigaMOT Challenge~\cite{panda}. The annotations are provided at 2 frames per second (FPS) and include pedestrian track IDs, bounding boxes, and face orientation labels, as shown in Fig.~\ref{fig:cat_pipeline}(a). The six CAT scenes cover diverse real-world environments, including university campuses, commercial areas, a primary school, and an urban intersection.

\subsection{Data Annotation and Processing}
\label{sec:annotations}

\paragraph{Pedestrian Attribute Annotations.}
Motivated by the behavioral differences discussed in Sec.~\ref{sec:intro}, we define four groups of pedestrian attributes that are visually identifiable and relevant to motions and interactions (Fig.~\ref{fig:cat_pipeline}(b)).

(1) \emph{Age} distinguishes between children and adults. We screen candidate pedestrians using bounding box sizes and body proportions, following prior work~\cite{ince2014child,kogure2022age}. We do not use facial features for age annotation. (2) \emph{Mobility mode} describes how individuals move, including movement on foot, use of personal mobility vehicles, and dependent mobility such as being carried or transported in a stroller. We use a YOLO-based model~\cite{yolo} to detect people and vehicles for candidate screening. (3) \emph{Accompanying children} indicates whether an adult pedestrian accompanies or cares for a child. Using age annotations, we identify candidate adults who remain within 1.2\,m of a child across multiple frames, based on Hall's proxemic zones~\cite{hall}. (4) \emph{Phone use} includes looking at a phone or holding it to the ear. We use each pedestrian’s face orientation relative to their walking direction to screen candidate phone users. 

After candidate screening, we manually review all pedestrians across multiple frames to correct annotations and identify missed cases. We follow annotation guidelines and use a web-based visualization tool that we developed for this task.

\vspace{-1em}
\paragraph{Homography Estimation.}
For each scene in the side-view videos, we compute a homography that maps image coordinates to world coordinates on the ground plane. This enables consistent measurement of pedestrian distances and speeds across diverse scenes.
We first select ground-level reference points in a video frame and obtain their corresponding world coordinates from Google Earth imagery, as illustrated in Fig.~\ref{fig:cat_pipeline}(c). Based on these correspondences, we estimate the homography $\mathbf{H}\in\mathbb{R}^{3\times3}$ using RANSAC~\cite{ransac}. We then apply this transformation to the bottom center of each pedestrian bounding box to reconstruct world-coordinate trajectories. Details of homography estimation and reprojection errors are provided in Appendix A.1.
% ~\ref{sec:supp_panda_homography}

\vspace{-1em} 
\paragraph{Scene Semantic Annotation.}
We construct scene semantic maps to provide spatial context for modeling pedestrian starting positions, attributes, goals, and motions. We use Grounded-SAM~\cite{grounded_sam} to segment scene regions and assign semantic labels. We annotate eight semantic classes: \emph{building}, \emph{structure}, \emph{bush}, \emph{grass}, \emph{tree}, \emph{sidewalk}, \emph{road}, and \emph{crosswalk}. The annotated regions are transformed into bird’s-eye-view (BEV) semantic maps using the estimated homographies (Fig.~\ref{fig:cat_pipeline}(d)). This aligns scene semantics with pedestrian trajectories in a common world coordinate system (Fig.~\ref{fig:cat_pipeline}(e)).

\begin{table}[t]
    \centering
    \input{tables/attribute_statistics}
\end{table}

\subsection{Attribute-dependent Behavior Analysis}
\label{sec:statistics}
We numerically analyze attribute-related behavioral differences in CAT in terms of kinematics, stopping behavior, and social proximity across attribute groups. Table~\ref{tab:attribute_statistics} summarizes the results. We compute speed from displacement over 1\,s intervals and consider a pedestrian stationary when their speed is below 0.3\,m/s. For moving periods, we report speed and the absolute rate of speed change (acceleration), and for stationary periods, we report their duration (stop time). Social proximity is measured by the distance to the nearest pedestrian (neighbor distance) and the proportion of moving frames with another pedestrian within 1.5\,m and a speed difference below 0.5\,m/s (companion rate). Compared with the \textit{Adult / on foot} group, children walk more slowly (0.90 vs.\ 1.09\,m/s) and stay closer to others (0.59 vs.\ 1.12\,m). Pedestrians accompanying children show patterns similar to those of children and a higher companion rate (63 vs.\ 37\%). Personal mobility users move faster (2.85\,m/s) and maintain greater distances from the nearest pedestrian (1.99\,m), whereas phone users stop for longer periods (7.6 vs.\ 3.5\,s) and have a lower companion rate (22\%). These group-specific behavioral differences remain consistent across scenes, as shown in Appendix A.3, motivating our crowd generation approach.

% ~\ref{sec:supp_scene_attr}

\begin{figure*}[t]
\centering
\includegraphics[width=\textwidth]{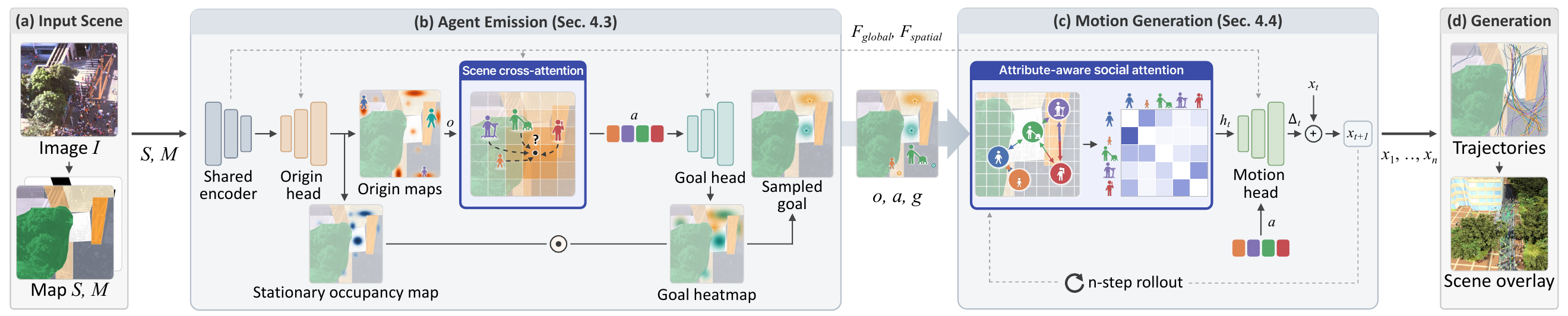}
\vspace{-6mm}
\caption{
\textbf{Overview of ACG.}
A shared U-Net encoder provides scene features to both agent emission and crowd motion.
Scene cross-attention predicts attribute distributions from agent origins and scene context.
Attribute-aware social attention models interactions based on attribute pairs and relative positions.
The sampled attributes condition goal generation and displacement prediction.
}
\vspace{-3mm}
\label{fig:model}
\end{figure*}

%% file: tables/attribute_statistics.tex
% Table: attribute-wise statistics (single-column)
\providecommand{\hd}[1]{}
\renewcommand{\hd}[1]{\begin{tabular}[c]{@{}c@{}}#1\end{tabular}}
\providecommand{\sd}[1]{{\scriptsize$\pm$#1}}
\providecommand{\iqr}[1]{{\scriptsize[#1]}}

\centering
\renewcommand{\arraystretch}{1.12}
\setlength{\tabcolsep}{2pt}

% \caption{
% \textbf{Attribute-level statistics in CAT.}
% Speed and acceleration are reported as mean\,$\pm$\,standard deviation,
% stop time and neighbor distance as median [Q1, Q3],
% and companion rate as a mean percentage.
% Speed: speed during movement;
% Acceleration: absolute rate of speed change during movement;
% Stop time: duration of stationary periods;
% Neighbor distance: distance to the nearest pedestrian;
% Companion rate: proportion of moving frames with another pedestrian
% within 1.5\,m and a speed difference below 0.5\,m/s.
% }
\caption{
\textbf{Attribute-level statistics in CAT.}
Speed and acceleration: mean $\pm$ standard deviation;
stop time and neighbor distance: median [Q1, Q3];
companion rate: mean percentage.
}
\label{tab:attribute_statistics}
\vspace{-2mm}

\resizebox{\linewidth}{!}{%
\begin{tabular}{lccccc}
\toprule
Feature
& \hd{Adult /\\[-2pt]on foot}
& Child
& \hd{Personal\\[-2pt]mobility}
& \hd{Accomp.\\[-2pt]a child}
& \hd{Phone\\[-2pt]use} \\
\midrule

\# Pedestrians
& 2,477 & 530 & 168 & 303 & 403 \\
\midrule

Speed (m/s)
& 1.09\,\sd{0.34}
& 0.90\,\sd{0.30}
& 2.85\,\sd{1.30}
& 0.87\,\sd{0.26}
& 1.05\,\sd{0.40} \\

Acceleration (m/s$^2$)
& 0.34\,\sd{0.26}
& 0.24\,\sd{0.16}
& 0.58\,\sd{0.36}
& 0.22\,\sd{0.14}
& 0.30\,\sd{0.22} \\

Stop time (s)
& 3.5\,\iqr{2.3, 9.4}
& 5.5\,\iqr{3.0, 13.9}
& 4.0\,\iqr{2.5, 11.1}
& 5.0\,\iqr{3.0, 13.5}
& 7.6\,\iqr{3.0, 25.2} \\

Neighbor dist.\ (m)
& 1.12\,\iqr{0.72, 1.85}
& 0.59\,\iqr{0.44, 1.04}
& 1.99\,\iqr{1.26, 3.26}
& 0.67\,\iqr{0.51, 1.09}
& 1.70\,\iqr{1.05, 2.29} \\

Companion (\%)
& 37 & 62 & 12 & 63 & 22 \\

\bottomrule
\end{tabular}%
}
\vspace{-3mm}

%% file: sec/4_method.tex
\section{Method}
\label{sec:method}

\subsection{Overview}
\label{sec:formulation}
Pedestrian attributes are related to both who appears in a scene and how they move and interact. We thus incorporate attributes into agent emission and motion generation to model scene-dependent pedestrian composition and attribute-dependent walking behavior.

Given a semantic map and a mask for traversable regions, ACG generates pedestrians with attributes and their trajectories over $T$ frames. As shown in \cref{fig:model}, the emission and motion modules share a scene encoder (\cref{sec:scene}). During generation, the emission module (\cref{sec:emission}) sequentially samples each agent's origin, attributes, and initial goal. The motion module (\cref{sec:sim}) uses these outputs as initial conditions and generates the agent's trajectory autoregressively, based on scene context and surrounding agents.

\subsection{Scene Representation}
\label{sec:scene}
Scene context is essential for determining pedestrian origins, attributes, and goals during emission and trajectories during motion generation. Since both modules depend on the same scene semantics and spatial layout, we use a shared encoder to extract global context and local spatial features.

The U-Net~\cite{unet} encoder $\theta_{\mathrm{enc}}$ takes the semantic map $S$, the mask $M$, and heatmaps $H_{\mathrm{ego}}$ and $H_{\mathrm{nbr}}$ representing recent positions of the target agent and its neighbors, respectively, as inputs. The encoder then extracts global scene features $F_{\mathrm{global}}$ and multi-scale spatial features $F_{\mathrm{spatial}}$:
\begin{equation}
(F_{\mathrm{global}},F_{\mathrm{spatial}})
=\theta_{\mathrm{enc}}([M,S,H_{\mathrm{ego}},H_{\mathrm{nbr}}]),
\label{eq:shared_encoder}
\end{equation}
where $[\cdot]$ denotes channel-wise concatenation. $F_{\mathrm{global}}$ captures the overall scene context, while $F_{\mathrm{spatial}}$ preserves local semantic and spatial details.

\subsection{Agent Emission}
\label{sec:emission}
The distribution of pedestrian attributes varies across scenes. To generate scene-dependent pedestrian composition, we condition attribute distributions on each agent's origin and scene context. Initial goals are then selected based on the origin and sampled attributes.

For each agent $i$, the emission module sequentially samples its origin $o^i$, attributes $a^i$, and initial goal $g^i$ as follows:
\begin{equation}
\begin{aligned}
o^i &\sim p_o(o\mid S,M),\\
a^i &\sim p_a(a\mid o^i,S,M),\\
g^i &\sim p_g(g\mid o^i,a^i,S,M).
\end{aligned}
\label{eq:emission}
\end{equation}
An origin head, a scene cross-attention module, and a goal head predict the distributions $p_o$, $p_a$, and $p_g$, respectively. Since an agent's position history is unavailable when it first appears in the scene, we set $H_{\mathrm{ego}}$ and $H_{\mathrm{nbr}}$ to zero, and compute scene features from the semantic map and the mask alone with the shared encoder.

\vspace{3mm}
\noindent \textbf{Origin head.}
To estimate where pedestrians appear in a scene, the origin head $\theta_{\mathrm{origin}}$ takes the shared scene features as input and outputs three types of logit maps:
\begin{equation}
(Z_{\mathrm{all}},Z_{\mathrm{attr}},Z_{\mathrm{occ}})
=\theta_{\mathrm{origin}}(F_{\mathrm{global}},F_{\mathrm{spatial}}).
\label{eq:origin_decoder}
\end{equation}
$Z_{\mathrm{all}}$ infers where pedestrians are likely to appear, while $Z_{\mathrm{attr}}$ provides attribute-specific origin maps for auxiliary supervision. $Z_{\mathrm{occ}}$ predicts where pedestrians stop or remain stationary, which guides goal sampling. We supervise these three outputs with binary cross-entropy against their corresponding target heatmaps.

To sample an agent's origin, we define the origin distribution by masking $Z_{\mathrm{all}}$ with $M$ to restrict origins to traversable regions and normalizing over the grid:
\begin{equation}
p_o(u\mid S,M)=\frac{\sigma(Z_{\mathrm{all}}(u))\,M(u)}{\sum_v \sigma(Z_{\mathrm{all}}(v))\,M(v)},
\label{eq:origin_distribution}
\end{equation}
where $\sigma$ is the sigmoid function and $u,v$ are grid-cell indices.\ We sample a grid location from $p_o$ and convert it to the corresponding world-coordinate origin $o^i$.

\vspace{3mm}
\noindent\textbf{Scene cross-attention.}
Pedestrian composition also differs across scenes. We model these differences by estimating attribute distributions from local features at each origin and global context, both extracted by \(\theta_{\mathrm{enc}}\). Here, we define $W_Q^{\mathrm{sc}}$, $W_K^{\mathrm{sc}}$, and $W_V^{\mathrm{sc}}$ as learnable projection matrices. For each origin $o^i$, we concatenate $F_{\mathrm{spatial}}(o^i)$ with its coordinates and project the result with $W_Q^{\mathrm{sc}}$ to form the query $Q_i^{\mathrm{sc}}$. We flatten the spatial dimensions of $F_{\mathrm{global}}$ into scene tokens and project them separately with $W_K^{\mathrm{sc}}$ and $W_V^{\mathrm{sc}}$ to obtain keys $K^{\mathrm{sc}}$ and values $V^{\mathrm{sc}}$. For each attention head, we compute:
\begin{equation}
O_i^{\mathrm{sc}}
=
\operatorname{softmax}\!\left(
\frac{Q_i^{\mathrm{sc}}(K^{\mathrm{sc}})^\top}
{\sqrt{d_{\mathrm{sc}}}}
\right)V^{\mathrm{sc}},
\label{eq:attribute_distribution}
\end{equation}
where $d_{\mathrm{sc}}$ is the query and key dimension of each head. The softmax normalizes attention weights over all scene tokens. We concatenate the outputs of all heads and apply an output projection. The result is passed to a separate softmax classifier for each of the four attribute groups defined in Sec.~\ref{sec:annotations}. The classifiers are supervised with categorical cross-entropy with annotated attributes at ground-truth origins. We sample one category from each predicted distribution and concatenate the corresponding one-hot encodings to form the attribute $a^i\sim p_a(a\mid o^i,S,M)$.

\vspace{3mm}
\noindent \textbf{Goal head.} 
The goal head predicts a heatmap of likely goals using scene features conditioned on pedestrian attributes. We first map \(a^i\) to an attribute embedding \(e^i=W_a{a^i}\) using a linear layer. We then use feature-wise linear modulation (FiLM)~\cite{perez2018film} to adjust $F_{\mathrm{global}}$ according to this embedding:

\begin{equation}
\widetilde{F}^i
=
\gamma(e^i)\odot F_{\mathrm{global}}
+
\beta(e^i),
\label{eq:film}
\end{equation}
where $\gamma$ and $\beta$ infer channel-wise scales and shifts from $e^i$. The goal head uses these features to produce the goal heatmap, which is supervised with binary cross-entropy between predicted goal heatmaps and ground-truth goal heatmaps. For goal sampling, we combine the predicted goal heatmap with \(Z_{\mathrm{occ}}\) and an origin-to-goal distance distribution estimated from training data.

\subsection{Motion Generation}
\label{sec:sim}
Pedestrians adjust their walking behaviors according to their surroundings, their own attributes, and those of nearby pedestrians. Using this information, we design the motion module to generate trajectories toward each pedestrian's goal after agent emission. At each step, the module predicts the next displacement from the pedestrian's recent motion and social context provided by the attribute-aware social attention.

\vspace{3mm}
\noindent\textbf{Social attention.}
We use attribute-aware social attention to model interactions based on relative positions and attributes. To incorporate this information into attention scores, we add learned biases. Prior work uses such biases to encode relative token positions~\cite{t5,swin} or graph structure~\cite{graphormer}. Let $r_t^{ij}$ denote the relative position vector from agent $i$ to agent $j$ in normalized coordinates at time $t$. We compute the bias:
\begin{equation}
B_t^{ij}
=
\phi_{\mathrm{rel}}(r_t^{ij})
+
\phi_{\mathrm{pair}}(e^i,e^j,r_t^{ij}),
\label{eq:social_bias}
\end{equation}
where $\phi_{\mathrm{rel}}$ and $\phi_{\mathrm{pair}}$ are separate MLPs that produce spatial and attribute-dependent biases, respectively. Thus, agent pairs at the same relative position can be assigned different biases depending on their attributes.

To compute attention weights at time $t$, we represent each agent as a token by combining learned representations of its current position and attributes. 
The agent tokens are stacked as rows to form the matrix $X_t$, with one row per agent.
Using learnable projection matrices $W_Q^{\mathrm{soc}}$, $W_K^{\mathrm{soc}}$, and $W_V^{\mathrm{soc}}$, we compute queries $Q_t^{\mathrm{soc}}=X_tW_Q^{\mathrm{soc}}$, keys $K_t^{\mathrm{soc}}=X_tW_K^{\mathrm{soc}}$, and values $V_t^{\mathrm{soc}}=X_tW_V^{\mathrm{soc}}$. For each attention head, we add $B_t$ to the query--key scores as follows:
\begin{equation}
O_t^{\mathrm{soc}}
=
\operatorname{softmax}\!\left(
\frac{Q_t^{\mathrm{soc}}(K_t^{\mathrm{soc}})^\top}
{\sqrt{d_{\mathrm{soc}}}}
+B_t
\right)V_t^{\mathrm{soc}},
\label{eq:social_attention}
\end{equation}
where $d_{\mathrm{soc}}$ is the query and key dimension of each head. Head indices are omitted throughout this formulation for simplicity. We concatenate the outputs of all heads and apply an output projection to obtain a social context representation for each agent. We also encode each agent's most recent $L$ velocity vectors, then combine this encoding with its social context representation using an MLP to produce the social feature $h_t^i$.

\vspace{-0.7em}
\paragraph{Displacement prediction.}
To predict each agent's next displacement, we concatenate its spatial feature $F_{\mathrm{spatial}}(x_t^i)$, social feature $h_t^i$, goal $g_t^i$, and current velocity $v_t^i$ to form the input vector $z_t^i$. The motion head \(\phi_{\mathrm{mot}}\) predicts the next-step displacement, with its output scaled as follows:
\begin{equation}
\Delta_t^i
=\exp(w^\top e^i)\,\phi_{\mathrm{mot}}(z_t^i,e^i),
\label{eq:motion_prediction}
\end{equation}
where $w$ is a learnable vector and $\exp(w^\top e^i)$ scales the displacement. $\phi_{\mathrm{mot}}$ conditions the concatenated features on the attribute embedding $e^i$ using FiLM, as in \cref{eq:film}. The resulting $\Delta_t^i$ is a one-step displacement in world coordinates.

\subsection{Generation and Training}

\paragraph{Generation.}
We sample scene population, group sizes, and agent lifetimes based on statistics computed from the training data. Agents are initialized according to \cref{eq:emission}. At each step, we add temporally correlated noise to the predicted displacements and then adjust them to avoid collisions and enforce agents' movement within traversable regions. Goals are resampled when reached, while attributes remain fixed. Agents are removed when they leave the scene or their lifetimes expire. Further details are provided in Appendix C.
% ~\ref{app:generation}

\vspace{-0.7em}
\paragraph{Training objectives.}
We jointly train the emission and motion modules through the shared scene encoder. The emission module uses a cross-entropy loss $\mathcal{L}_{\mathrm{ce}}$, while the motion module uses a trajectory reconstruction loss $\mathcal{L}_{\mathrm{traj}}$ and a behavior matching loss $\mathcal{L}_{\mathrm{beh}}$:
\begin{equation}
\mathcal{L}_G
=
\mathcal{L}_{\mathrm{ce}}
+
\mathcal{L}_{\mathrm{traj}}
+
\mathcal{L}_{\mathrm{beh}}.
\label{eq:generator_objective}
\end{equation}
$\mathcal{L}_{\mathrm{ce}}$ combines binary cross-entropy for origin, stationary occupancy, and goal maps with categorical cross-entropy for attribute prediction.
$\mathcal{L}_{\mathrm{traj}}$ combines an $\ell_1$ position loss and soft-DTW~\cite{cuturi2017soft} to match generated and observed trajectories.
$\mathcal{L}_{\mathrm{beh}}$ encourages attribute-dependent motion and realistic pedestrian spacing through three components:
\begin{equation}
\mathcal{L}_{\mathrm{beh}}
=
\lambda_{\mathrm{adv}}\mathcal{L}_{\mathrm{adv}}
+
\lambda_{\mathrm{spd}}\mathcal{L}_{\mathrm{spd}}
+
\lambda_{\mathrm{nn}}\mathcal{L}_{\mathrm{nn}}.
\label{eq:beh}
\end{equation}
The adversarial loss $\mathcal{L}_{\mathrm{adv}}$ uses an attribute-conditioned LSGAN discriminator~\cite{mao2017least} to encourage realistic velocity sequences. The speed loss $\mathcal{L}_{\mathrm{spd}}$ matches quantiles of generated and observed speed distributions for each attribute value. The spacing loss $\mathcal{L}_{\mathrm{nn}}$ matches nearest-neighbor distance distributions between generated and observed crowds. 
% We set $\lambda_{\mathrm{adv}}=0.3$, $\lambda_{\mathrm{spd}}=0.5$, and $\lambda_{\mathrm{nn}}=0.3$ in all experiments. 
Detailed loss definitions, weights, and the discriminator objective are provided in Appendix D.

% ~\ref{app:losses}

%% file: sec/5_experiments.tex
\section{Experiments}
\label{sec:exp}

\subsection{Setup}
\paragraph{Datasets and evaluation protocol.}
We evaluate ACG on ETH~\cite{eth}/UCY~\cite{ucy}, SDD~\cite{sdd}, and GCS~\cite{GCS,gcs2}, established public benchmarks for human trajectory prediction.
For CAT, we perform two-fold temporal cross-validation on six scenes, swapping the training and test segments to obtain twelve evaluations. Additionally, to demonstrate downstream effectiveness, we conduct social navigation experiments on CAT.

\vspace{-0.5em}
\paragraph{Metrics.}
To assess how closely generated crowds reproduce real crowd statistics and pedestrian trajectories, we adopt the scene-level realism and agent-level similarity metrics from CrowdES~\cite{crowdes}. At the scene level, density (Dens.) measures the mean number of agents per grid cell. Frequency (Freq.) measures the mean number of distinct agent types per cell. Coverage (Cov.) measures the fraction of occupied cells, while population (Pop.) measures the number of agents in the scene. We compute these statistics at each sampled time step and compare the generated and real distributions over time using Earth Mover's Distance (EMD)~\cite{emd}.

At the agent level, kinematics (Kin.) averages the EMDs for speed, acceleration, travel distance, and duration distributions. Dynamic Time Warping (DTW)~\cite{dtw} averages the mean nearest-trajectory DTW distances in the generated-to-real and real-to-generated directions. Diversity (Div.) measures trajectory-set coverage under bidirectional nearest-track matching, while collision rate (Col.) quantifies agent--agent collisions during simulation.

\subsection{Quantitative Results}
\paragraph{CAT Benchmark.}
We compare ACG with the algorithmic baseline SE-ORCA~\cite{orca} and the learning-based methods VAE~\cite{pecnet}, TRACE~\cite{trace}, and CrowdES~\cite{crowdes}.

\cref{tab:main} shows that ACG achieves the best or tied-best results on seven of eight metrics. Compared with CrowdES, ACG reduces errors in frequency, a measure of local agent-type diversity, as well as kinematic statistics and DTW by 28.9\%, 43.5\%, and 34.3\%, respectively. These improvements indicate a closer match to local agent-type diversity and observed motion patterns, although CrowdES better matches the distribution of spatial coverage. SE-ORCA achieves a low collision rate through explicit collision-avoidance constraints. ACG matches its collision rate of 0.001\% while generating more diverse trajectories (0.340 vs.\ 0.250) and better reproducing scene-level crowd statistics. Overall, ACG reproduces diverse trajectory patterns and realistic motion while maintaining a low collision rate.

\input{tables/CAT_benchmark}
\input{tables/public_benchmark}
\input{tables/latency}

\vspace{-0.5em}
\paragraph{Public Benchmarks.}
\cref{tab:public} shows that ACG achieves the best results on six ETH/UCY metrics and best or tied-best results on four SDD metrics, with second-best results on the remaining metrics in both datasets. These results support its ability to reproduce scene-level crowd statistics and agent-level motion patterns. On GCS, ACG achieves the best results on all four scene-level metrics and kinematic statistics. Compared with TRACE, ACG has higher DTW and lower diversity but improves the other six metrics, reducing all four scene-level errors by 52.2--66.3\% and the collision rate by 23.5\%. These results extend the evidence of ACG's effectiveness beyond CAT to diverse public benchmarks.

\vspace{-1em}
\paragraph{Computational Efficiency.}
To evaluate computational efficiency, we measure runtime per agent-step. As shown in \cref{tab:latency}, ACG achieves the lowest runtime at 1.81\,ms per agent-step, compared with 2.20\,ms for CrowdES. Although TRACE performs well on public benchmarks, it requires 96.9\,ms per agent-step, approximately 54 times ACG's runtime. Our model uses lightweight modules to generate trajectories without iterative denoising. The benchmark results and runtime measurements show that ACG combines strong crowd generation performance with low computational cost.

\input{tables/attribute_fidelity}

\input{tables/attribute_ablation}

\subsection{Attribute Analysis}
\label{sec:attribute_fidelity}
\begin{figure}[t]
    \centering
    \includegraphics[width=\columnwidth]{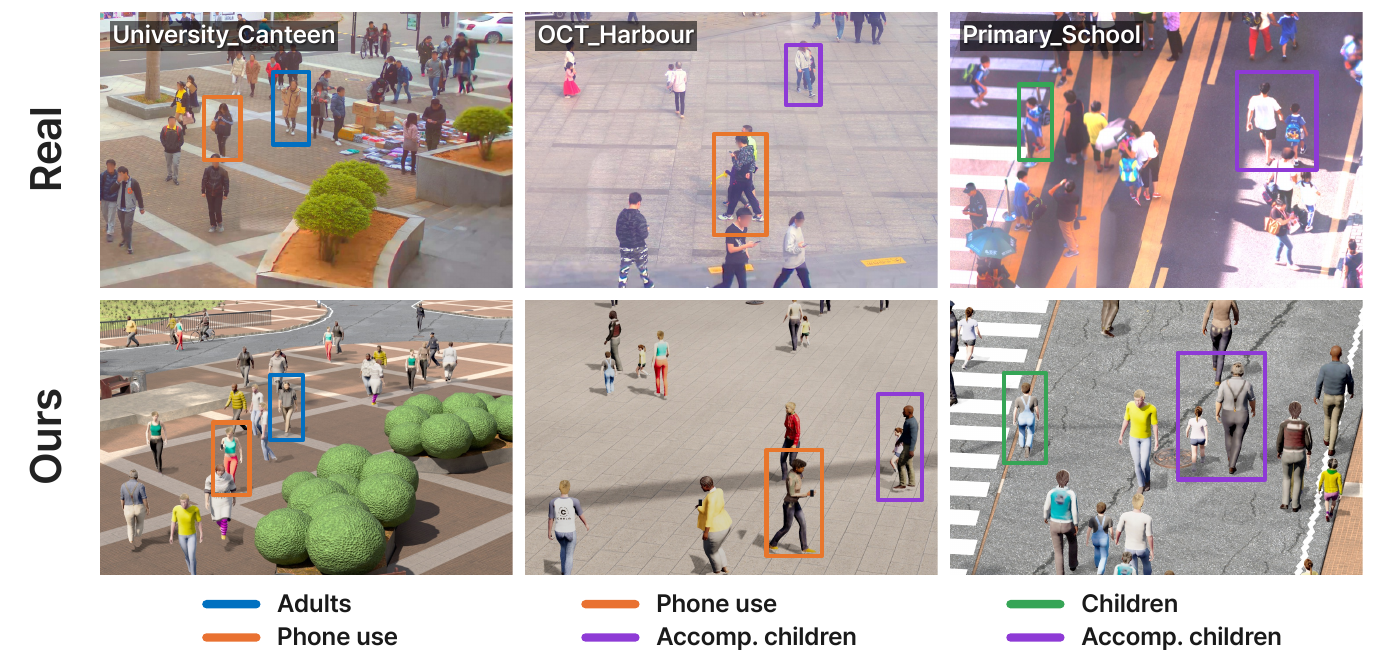}
    \vspace{-6mm}
    \caption{\textbf{Qualitative comparison of real and generated crowds across scenes.} Real crowds (top) and ACG-generated crowds (bottom) illustrate scene-specific pedestrian composition. Colored boxes highlight examples of attributes prevalent in each scene.}
\label{fig:acg_comp}
\vspace{-3mm}
\end{figure}

We evaluate how closely generated crowds match real pedestrian composition and attribute-specific mean speeds in \cref{tab:evalv}. ACG achieves smaller absolute errors than CrowdES for every reported agent proportion, indicating a closer match to the ground-truth pedestrian composition. ACG also better reproduces all attribute-specific mean speeds. For personal mobility, ACG generates a proportion of 5.0\% and a mean speed of 2.06\,m/s, close to the real values of 4.5\% and 2.23\,m/s, whereas CrowdES underestimates both at 2.6\% and 0.39\,m/s. \cref{fig:acg_comp} further shows visualizations of generated crowds in CARLA~\cite{carla}, built on Unreal Engine~\cite{unrealsoftware}. These visualizations demonstrate scene-specific pedestrian composition, including phone users near the university and children with accompanying adults near the primary school.

\subsection{Ablation Study}
\label{sec:ablation}
We evaluate the effectiveness of scene cross-attention and attribute-aware social attention by removing each module separately and both simultaneously, with results reported in \cref{tab:ablation_metrics}. Removing scene cross-attention increases frequency error from 0.064 to 0.079, indicating its contribution to reproducing local agent-type diversity. Removing attribute-aware social attention increases the kinematic error from 0.343 to 0.350 and DTW from 24.62 to 25.06, indicating its contribution to capturing the ground-truth motion patterns. Although some ablations perform slightly better on individual metrics, the results support the effectiveness of both modules in modeling pedestrian composition and motion.

\input{tables/navigation}
\input{tables/user_study}

\subsection{Navigation Policy Evaluation}
\label{sec:nav}
To assess the effectiveness of our crowds for robot navigation training, we conduct simulation experiments in CAT scenes. We train SARL~\cite{SARL}, RGL~\cite{RGL}, LSTM-RL~\cite{lstm-rl}, and CrowdNav++~\cite{crowdnav} on crowds generated by either CrowdES or ACG, then evaluate each policy on both generators' crowds in unseen CAT scenes. Following~\cite{SARL,RGL}, we report success rate, the proportion of evaluation episodes in which the robot reaches its goal without collision, and collision rate, the proportion involving a robot--pedestrian collision.

\begin{figure}[t]
    \centering
    \includegraphics[width=\columnwidth]{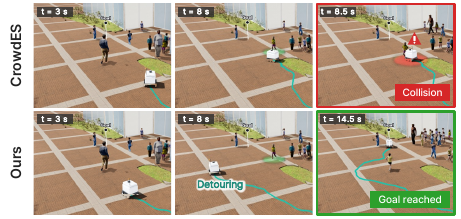}
    \vspace{-6mm}
    \caption{\textbf{Qualitative comparison of robot navigation.} The CrowdES-trained policy (top) collides with a child at $t{=}8.5$\,s, whereas the ACG-trained policy (bottom) detours around the children and reaches the goal at $t{=}14.5$\,s.}

\label{fig:nav_qual}
\vspace{-3mm}
\end{figure}

As shown in \cref{tab:nav}, training on ACG-generated crowds reduces collision rates across all four policies under both generators. While success rates decrease for CrowdNav++ and for SARL on CrowdES crowds, RGL and LSTM-RL achieve higher success rates under both generators. For example, training RGL on ACG crowds reduces its collision rate from 5.1\% to 2.1\% and increases its success rate from 93.8\% to 96.6\% when evaluated on CrowdES crowds. \cref{fig:nav_qual} illustrates an ACG-trained policy safely detouring around children. These results show that ACG-generated crowds support effective collision-avoidance training, with improvements extending to crowds generated by another model.

% To assess the benefits of our crowds for robot navigation training, we train SARL~\cite{SARL}, RGL~\cite{RGL}, LSTM-RL~\cite{lstm-rl}, and CrowdNav++~\cite{crowdnav} on crowds generated by either CrowdES or ACG. Each trained policy is evaluated on crowds generated by both models in unseen scenes.

% As shown in \cref{tab:nav}, training on crowds generated by ACG lowers collision rates across all four navigation policies under both evaluation generators, suggesting benefits beyond the training generator. For SARL evaluated on ACG crowds, collision rates decrease from 12.0\% to 7.4\%, while success rates increase from 85.2\% to 89.2\%. RGL and LSTM-RL also achieve higher success rates under both generators. However, success rates decrease for CrowdNav++ under both generators and for SARL on CrowdES crowds, showing that improved collision avoidance does not always translate into higher goal-reaching success. \cref{fig:nav_qual} illustrates an ACG-trained policy safely navigating a scene with children by taking a wider detour around them.

\subsection{User Study}
\label{sec:user_study}
To assess perceived crowd realism, scene–attribute consistency, attribute–motion consistency, and perceived navigation safety, we conduct a user study with 50 participants recruited through CloudResearch~\cite{hartman2023introducing}. We collect 900 responses across 18 video pairs. As shown in \cref{tab:user_study}, participants prefer ACG crowds and policies trained on them over their CrowdES counterparts across all criteria. Preference rates reach 88.0\% for composition, 86.0\% for attribute--motion consistency, and 91.8\% for navigation safety.

% We conduct a user study with 50 participants recruited through CloudResearch~\cite{hartman2023introducing}, collecting 900 responses across 18 video pairs. Participants assess perceived crowd realism, the appropriateness of pedestrian attributes for each scene, and perceived navigation safety. Participants evaluate movement realism, scene-dependent composition, attribute--motion consistency, and navigation safety. As shown in \cref{tab:user_study}, participants prefer ACG crowds and policies trained on them over CrowdES counterparts across all criteria. Preferences reach 88.0\% for composition, 86.0\% for attribute--motion consistency, and 91.8\% for navigation safety.

%% file: tables/CAT_benchmark.tex
\begin{table}[t]
\centering
\footnotesize
\renewcommand{\arraystretch}{1.12}
\setlength{\tabcolsep}{2.5pt}

\caption{
\textbf{Crowd generation on CAT.}
Best and second-best results are bold and underlined, respectively.
}
\label{tab:main}
\vspace{-2mm}

\resizebox{\columnwidth}{!}{%
\begin{tabular}{lcccccccc}
\toprule
Method
& Dens$\downarrow$ & Freq$\downarrow$
& Cov$\downarrow$ & Pop$\downarrow$
& Kin$\downarrow$ & DTW$\downarrow$
& Div$\uparrow$ & Col$\downarrow$ \\
\midrule

SE-ORCA
& 0.553 & 0.482 & 0.359 & 0.167
& \underline{0.400} & 49.27 & 0.250 & \textbf{0.001} \\

VAE
& 0.486 & 0.092 & 0.074 & 0.160
& 0.607 & 37.68 & \underline{0.253} & 0.024 \\

TRACE
& \underline{0.422} & 0.124 & 0.058 & \underline{0.148}
& 0.775 & 44.00 & 0.229 & \underline{0.017} \\

CrowdES
& \underline{0.422} & \underline{0.090}
& \textbf{0.040} & 0.154
& 0.607 & \underline{37.48} & 0.251 & 0.025 \\

\rowcolor{oursgray}
\textbf{ACG (ours)}
& \textbf{0.327} & \textbf{0.064}
& \underline{0.057} & \textbf{0.117}
& \textbf{0.343} & \textbf{24.62}
& \textbf{0.340} & \textbf{0.001} \\

\bottomrule
\end{tabular}%
}
\vspace{-3mm}
\end{table}

% \begin{table}[t]
% \centering
% \footnotesize
% \renewcommand{\arraystretch}{1.12}
% \setlength{\tabcolsep}{2.5pt}
% \resizebox{\columnwidth}{!}{%
% \begin{tabular}{lcccccccc}
% \toprule
% Method
% & Dens$\downarrow$ & Freq$\downarrow$
% & Cov$\downarrow$ & Pop$\downarrow$
% & Kin$\downarrow$ & DTW$\downarrow$
% & Div$\uparrow$ & Col$\downarrow$ \\
% \midrule

% SE-ORCA
% & 0.553 & 0.482 & 0.359 & 0.167
% & \underline{0.400} & 49.27 & 0.250 & \textbf{0.001} \\

% VAE
% & 0.486 & 0.092 & 0.074 & 0.160
% & 0.607 & 37.68 & \underline{0.253} & \underline{0.024} \\

% CrowdES
% & \underline{0.422} & \underline{0.090}
% & \textbf{0.040} & \underline{0.154}
% & 0.607 & \underline{37.48} & 0.251 & 0.025 \\

% \rowcolor{oursgray}
% \textbf{ACG (ours)}
% & \textbf{0.327} & \textbf{0.064}
% & \underline{0.057} & \textbf{0.117}
% & \textbf{0.343} & \textbf{24.62}
% & \textbf{0.340} & \textbf{0.001} \\

% \bottomrule
% \end{tabular}%
% }
% \caption{
% \textbf{Crowd generation on CAT.}
% Best and second-best results are bold and underlined, respectively.
% }
% \label{tab:main}
% \end{table}

%% file: tables/public_benchmark.tex
\begin{table}[t]
\centering
\footnotesize
\renewcommand{\arraystretch}{1.12}
\setlength{\tabcolsep}{1.5pt}

\caption{
\textbf{Crowd generation on ETH/UCY, SDD, and GCS.}
ETH/UCY results are averaged over five scenes.
Unmarked ETH/UCY and SDD baseline results are from~\cite{crowdes}.
GCS uses a common test split constructed for this study.
$\dagger$ denotes reimplemented baselines.
\textbf{Bold}/\underline{underline} indicate best/second-best results.
}
\label{tab:public}
\vspace{-2mm}

\resizebox{\columnwidth}{!}{%
\begin{tabular}{llcccccccc}
\toprule
Dataset & Method
& Dens$\downarrow$ & Freq$\downarrow$
& Cov$\downarrow$ & Pop$\downarrow$
& Kin$\downarrow$ & DTW$\downarrow$
& Div$\uparrow$ & Col$\downarrow$ \\
\midrule

ETH / & SE-ORCA
& 0.130 & 0.086 & 0.086 & 0.670
& 0.495 & 1.310 & 0.243 & \underline{0.025} \\

UCY & VAE
& 0.132 & 0.086 & 0.086 & 0.670
& 0.690 & 1.676 & 0.274 & 1.737 \\

& TRACE$^\dagger$
& 0.089 & \underline{0.053}
& \underline{0.053} & 0.315
& 1.095 & 0.988
& \textbf{0.310} & 1.221 \\

& CrowdES
& \underline{0.081} & \underline{0.053}
& \underline{0.053} & \underline{0.283}
& \textbf{0.337} & \underline{0.933}
& 0.289 & 0.916 \\

\rowcolor{oursgray}
\cellcolor{white} & \textbf{ACG (ours)}
& \textbf{0.057} & \textbf{0.031}
& \textbf{0.031} & \textbf{0.191}
& \underline{0.418} & \textbf{0.912}
& \underline{0.299} & \textbf{0.005} \\

\midrule

SDD & SE-ORCA
& 0.058 & 0.047 & \underline{0.044} & 0.540
& 0.893 & 6.645 & 0.377 & \underline{0.048} \\

& VAE
& 0.052 & 0.052 & 0.047 & 0.678
& 1.078 & 7.881 & \textbf{0.387} & 2.470 \\

& TRACE$^\dagger$
& 0.044 & \textbf{0.031}
& \textbf{0.030} & \textbf{0.430}
& 0.909 & \textbf{5.519}
& 0.346 & 0.161 \\

& CrowdES
& \textbf{0.038} & \underline{0.033}
& \textbf{0.030} & 0.463
& \underline{0.650} & 6.352
& 0.354 & 0.411 \\

\rowcolor{oursgray}
\cellcolor{white} & \textbf{ACG (ours)}
& \underline{0.040} & \textbf{0.031}
& \textbf{0.030} & \underline{0.459}
& \textbf{0.631} & \underline{5.845}
& \underline{0.380} & \textbf{0.000} \\

\midrule

GCS$^\dagger$ & SE-ORCA
& 1.099 & 0.195 & 0.195 & 2.539
& 1.088 & 3.212 & 0.244 & \textbf{0.002} \\

& VAE
& 0.978 & 0.376 & 0.376 & 2.259
& 1.040 & 3.836 & 0.319 & 10.624 \\

& TRACE
& 0.686 & 0.341 & 0.341 & 1.584
& \underline{0.401} & \textbf{2.483}
& \textbf{0.334} & 3.492 \\

& CrowdES
& \underline{0.332} & \underline{0.121}
& \underline{0.121} & \underline{0.768}
& 1.192 & 5.064 & 0.316 & 10.279 \\

\rowcolor{oursgray}
\cellcolor{white} & \textbf{ACG (ours)}
& \textbf{0.328} & \textbf{0.115}
& \textbf{0.115} & \textbf{0.757}
& \textbf{0.393} & \underline{2.825}
& \underline{0.320} & \underline{2.672} \\

\bottomrule
\end{tabular}%
}
\vspace{-3mm}
\end{table}

% GCS & SE-ORCA
% & 1.441 & \textbf{0.066}
% & \textbf{0.066} & \underline{0.617}
% & 3.419 & \underline{3.289}
% & 0.197 & \textbf{0.003} \\

% & VAE
% & 0.826 & \underline{0.085}
% & \underline{0.085} & \textbf{0.433}
% & 5.529 & 5.351 & 0.227 & 0.090 \\

% & CrowdES
% & \underline{0.332} & 0.121
% & 0.121 & 0.768
% & \underline{1.192} & 5.064
% & \textbf{0.334} & 0.341 \\

% \rowcolor{oursgray}
% \cellcolor{white} & \textbf{ACG (ours)}
% & \textbf{0.327} & 0.115
% & 0.115 & 0.755
% & \textbf{0.488} & \textbf{3.001}
% & \underline{0.309} & \underline{0.068} \\

%% file: tables/latency.tex
\begin{table}[t]
\centering
\small
\renewcommand{\arraystretch}{1.1}
\setlength{\tabcolsep}{4pt}

\caption{\textbf{Runtime per agent-step.} \textbf{Bold} indicates the lowest time.}
\label{tab:latency}
\vspace{-2mm}

\resizebox{\columnwidth}{!}{%
\begin{tabular}{lccccc}
\toprule
Method & ORCA & VAE & TRACE & CrowdES & \cellcolor{oursgray}\textbf{ACG (ours)} \\
\midrule
Time (ms/agent-step)$\downarrow$
& 1.97 & 24.20 & 96.90 & 2.20 & \cellcolor{oursgray}\textbf{1.81} \\
\bottomrule
\end{tabular}%
}
\vspace{-3mm}
\end{table}

%% file: tables/attribute_fidelity.tex
\begin{table}[t]
\centering
\footnotesize
\renewcommand{\arraystretch}{1.12}
\setlength{\tabcolsep}{2.5pt}

\caption{
\textbf{Attribute fidelity on CAT.}
Proportions and mean speeds are pooled over 12 evaluations.
PM: personal mobility; DM: dependent mobility;
Accomp: accompanying children; Phone: phone use. \textbf{Bold} indicates the smaller absolute error in each comparison.
}
\label{tab:evalv}
\vspace{-2mm}

\resizebox{\columnwidth}{!}{%
\begin{tabular}{lcc>{\columncolor{oursgray}}ccc>{\columncolor{oursgray}}c}
\toprule
& \multicolumn{3}{c}{Agent proportion}
& \multicolumn{3}{c}{Mean speed (m/s)} \\
\cmidrule(l){2-4}
\cmidrule(l){5-7}
Attribute
& Real & CrowdES & \textbf{ACG (ours)}
& Real & CrowdES & \textbf{ACG (ours)} \\
\midrule

Adult
& 0.938
& 0.972
& \textbf{0.942}
& 0.88 & 0.98 & \textbf{0.94} \\

Child
& 0.154 & 0.097 & \textbf{0.133}
& 0.58 & 0.82 & \textbf{0.79} \\

PM
& 0.045 & 0.026 & \textbf{0.050}
& 2.23 & 0.39 & \textbf{2.06} \\

DM
& 0.017 & 0.002 & \textbf{0.007}
& 0.69 & 0.21 & \textbf{0.87} \\

Accomp
& 0.099 & 0.022 & \textbf{0.057}
& 0.62 & 0.90 & \textbf{0.82} \\

Phone
& 0.119 & 0.092 & \textbf{0.125}
& 0.83 & 1.22 & \textbf{1.15} \\

\bottomrule
\end{tabular}%
}
\vspace{-3mm}
\end{table}

%% file: tables/attribute_ablation.tex
\begin{table}[t]
\centering
\footnotesize
\renewcommand{\arraystretch}{1.12}
\setlength{\tabcolsep}{2.5pt}

\caption{
\textbf{Component ablation on CAT.}
\textbf{Bold}/\underline{underline} indicate best/second-best results.
}
\label{tab:ablation_metrics}
\vspace{-2mm}

\resizebox{\columnwidth}{!}{%
\begin{tabular}{lcccccccc}
\toprule
Variant
& Dens$\downarrow$ & Freq$\downarrow$
& Cov$\downarrow$ & Pop$\downarrow$
& Kin$\downarrow$ & DTW$\downarrow$
& Div$\uparrow$ & Col$\downarrow$ \\
\midrule

w/o both attn.
& 0.325 & 0.086
& \underline{0.062} & \underline{0.118}
& \underline{0.333} & \underline{24.62}
& 0.339 & \underline{0.001} \\

w/o scene attn.
& \textbf{0.322} & 0.079
& \underline{0.062} & 0.119
& \textbf{0.331} & 24.83
& \textbf{0.342} & 0.002 \\

w/o social attn.
& \underline{0.324} & \underline{0.070}
& 0.066 & 0.123
& 0.350 & 25.06
& \underline{0.340} & \underline{0.001} \\
\rowcolor{oursgray}
\textbf{ACG (ours)}
& 0.327 & \textbf{0.064}
& \textbf{0.057} & \textbf{0.117}
& 0.343 & \textbf{24.62}
& \underline{0.340} & \textbf{0.001} \\

\bottomrule
\end{tabular}%
}
\vspace{-3mm}
\end{table}

%% file: tables/navigation.tex
\begin{table}[t]
\centering

\caption{
\textbf{Navigation policy evaluation.}
Policies trained on CrowdES or ACG crowds are cross-tested on
generated crowds.
}
\label{tab:nav}
\vspace{-2mm}

\resizebox{\columnwidth}{!}{%
\begin{tabular}{llcccc}
\toprule
& & \multicolumn{2}{c}{CrowdES crowds} & \multicolumn{2}{c}{ACG crowds} \\
\cmidrule(lr){3-4}\cmidrule(lr){5-6}
Policy & Simulator & Success$\uparrow$ & Collision$\downarrow$ & Success$\uparrow$ & Collision$\downarrow$ \\
\midrule
SARL & CrowdES & \textbf{0.964} & 0.022 & 0.852 & 0.120 \\
\rowcolor{oursgray}
\cellcolor{white} & \textbf{ACG (ours)} & 0.956 & \textbf{0.018} & \textbf{0.892} & \textbf{0.074} \\
\midrule
RGL & CrowdES & 0.938 & 0.051 & 0.852 & 0.127 \\
\rowcolor{oursgray}
\cellcolor{white} & \textbf{ACG (ours)} & \textbf{0.966} & \textbf{0.021} & \textbf{0.908} & \textbf{0.074} \\
\midrule
LSTM-RL & CrowdES & 0.967 & 0.024 & 0.865 & 0.113 \\
\rowcolor{oursgray}
\cellcolor{white} & \textbf{ACG (ours)} & \textbf{0.973} & \textbf{0.017} & \textbf{0.890} & \textbf{0.092} \\
\midrule
CrowdNav++ & CrowdES & \textbf{0.611} & 0.090 & \textbf{0.491} & 0.245 \\
\rowcolor{oursgray}
\cellcolor{white} & \textbf{ACG (ours)} & 0.553 & \textbf{0.064} & 0.485 & \textbf{0.157} \\
\bottomrule
\end{tabular}%
}
\vspace{-3mm}
\end{table}

%% file: tables/user_study.tex
\begin{table}[t]
    \centering
    \small
    \setlength{\tabcolsep}{2pt}
    \renewcommand{\arraystretch}{1.15}

    \caption{\textbf{User study.} We report preference percentages for ACG and CrowdES. Tie denotes no preference.}
    \label{tab:user_study}
    \vspace{-2mm}

    \begin{tabularx}{\columnwidth}{l*{4}{>{\centering\arraybackslash}X}}
        \toprule
        & \multicolumn{3}{c}{\textbf{Crowd generation}} & \textbf{Robot} \\
        \cmidrule(lr){2-4} \cmidrule(l){5-5}
        Preference & Realism & Composition & Attribute & Safety \\
        \midrule
        CrowdES & 3.3\% & 8.0\% & 6.0\% & 2.2\% \\
        Tie & 13.3\% & 4.0\% & 8.0\% & 6.0\% \\
        \rowcolor{oursgray}
        \textbf{ACG (ours)} & \textbf{83.3\%} & \textbf{88.0\%} & \textbf{86.0\%} & \textbf{91.8\%} \\
        \bottomrule
    \end{tabularx}
    \vspace{-3mm}
\end{table}

%% file: sec/6_conclusion.tex
\section{Conclusion}
To the best of our knowledge, we introduce the first attribute-aware crowd generation pipeline. We construct CAT, a dataset combining world-coordinate trajectories, semantic maps, and attribute annotations. Building on CAT, we propose ACG, which models scene-dependent composition and attribute-dependent behavior through a scene cross-attention and an attribute-aware social attention. Experiments demonstrate that our model reproduces real-world crowd statistics and trajectory patterns across CAT and public benchmarks. Navigation policies trained on our crowds achieve lower collision rates with comparable success rates.

\vspace{-0.5em}
\paragraph{Limitations and Future Work.}
Our model focuses on pedestrian trajectories and does not capture physical contact or high-level social interactions such as group formation and conversation. Modeling physical contact requires integrating human reconstruction with contact models in physics simulators, while modeling social interactions requires detailed annotations of group relationships and interaction types. Future work will extend our approach to cover these physical and social interactions for more realistic crowd simulation.

%% file: sec/X_suppl.tex
\newcommand{\supptitlefigure}{%
    \begin{minipage}{\textwidth}
        \centering
        \captionsetup{type=figure}
        \includegraphics[
            width=\linewidth,
            height=0.28\textheight,
            keepaspectratio
        ]{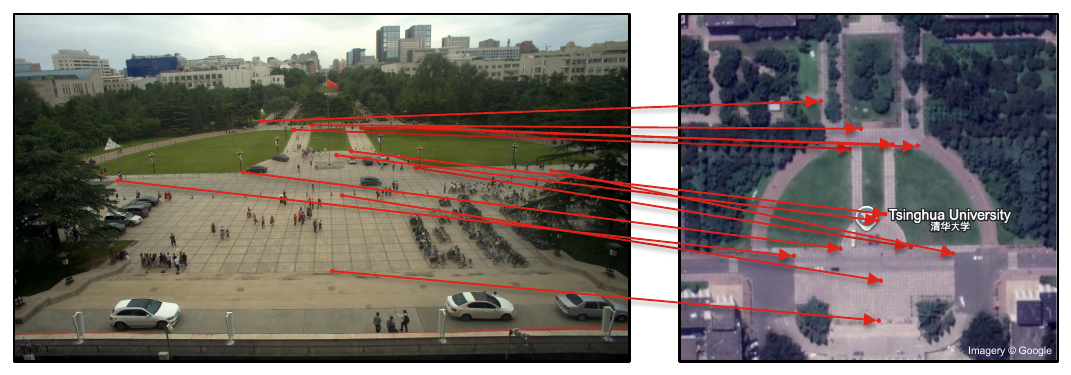}
        \caption{\textbf{Image-to-world mapping.} Pixel coordinates (left) and corresponding world coordinates (right).}
        \label{fig:reference_point_selection}
        \par\vspace{0.7em}
    \end{minipage}%
}

\patchcmd{\maketitlesupplementary}{\vspace{1.0em}}{\vspace{1.0em}\normalsize\supptitlefigure}{}{\PackageError{suppl}{title patch failed}{}}

\maketitlesupplementary
\appendix
\suppressfloats[t]

This supplementary material provides additional details on our dataset, method, and evaluation.
We describe how we construct our dataset and present scene-wise dataset statistics in \cref{app:cat_details}.
We explain the model architecture in \cref{app:architecture}, the crowd generation procedure in \cref{app:generation}, and the training procedure and loss functions in \cref{app:losses}.
We compare generated and real trajectories in learned feature spaces in \cref{app:feature_comparison} and report runtime measurements in \cref{app:runtime}.
Finally, we present navigation policy evaluation for different pedestrian attributes in \cref{app:nav_attribute} and describe the user study in \cref{app:user_study}.

\section{CAT Dataset Construction and Analysis}
\label{app:cat_details}

\cref{sec:supp_panda_homography} explains how we estimate homographies to obtain world-coordinate pedestrian trajectories and assess the results through reprojection errors and speed statistics.
\cref{sec:supp_cat_construction} provides an overview of the dataset construction process, including scene selection, pedestrian attribute annotation, and scene semantic map construction, and summarizes the final dataset statistics.
\cref{sec:supp_scene_attr} uses scene-wise statistics and visualizations to examine whether the attribute-dependent motion patterns observed in the pooled data also appear within individual scenes.

\subsection{Homography Estimation Details}
\label{sec:supp_panda_homography}

For each scene, we estimate a homography from ground points manually matched between the video frames and Google Earth imagery. We use this mapping to reconstruct pedestrian trajectories in world coordinates. We assess how closely the projected points match their Google Earth reference positions and examine pedestrian speeds to check the spatial scale of the reconstructed trajectories.

\paragraph{Reference Point Selection.}
For each scene, we select ground points visible in both the video frame and the corresponding Google Earth imagery. We use recognizable features such as road-marking corners, distinct points along curbs, and points where building walls meet the ground. We distribute these points across the walkable region and avoid elevated or non-planar features such as rooftops, tree canopies, and stairs. We choose points that are spread out across the image and avoid placing too many in one area or along the same straight line. For each reference point, we first mark its location in Google Earth and then select the same location in the gigapixel video frame. We use the same point ID $i$ in both images to pair the original pixel coordinates $(u_i,v_i)$ with the latitude and longitude $(\phi_i,\lambda_i)$ recorded in Google Earth. \cref{fig:reference_point_selection} shows an example of the ground points manually matched between a video frame and Google Earth imagery.

\paragraph{Latitude and Longitude to World Coordinates.}
Google Earth provides the location of each reference point as latitude and longitude in degrees. To measure distances in meters, we convert these locations to world coordinates. For each scene, we use the reference point with the smallest shared ID as the origin. Its geographic coordinates are $(\phi_0,\lambda_0)$, and its world coordinates are $(0,0)$. For a reference point $i$ with latitude $\phi_i$ and longitude $\lambda_i$, we compute
\begin{equation}
\begin{aligned}
E_i &= (\lambda_i-\lambda_0)\,s_\lambda,\\
N_i &= (\phi_i-\phi_0)\,s_\phi.
\end{aligned}
\label{eq:supp_local_world}
\end{equation}
Here, $s_\lambda$ and $s_\phi$ give the approximate lengths in meters of one degree of longitude and latitude, respectively, at the scene origin's latitude $\phi_0$. We compute these factors for each scene using the World Geodetic System 1984 (WGS~84) ellipsoid~\cite{united1987department} and apply them in \cref{eq:supp_local_world} to obtain the local coordinates $(E_i,N_i)$ in meters. A positive $E_i$ indicates that the point lies east of the origin, and a positive $N_i$ indicates that it lies north.

\paragraph{Homography Estimation.}
We assume that the reference points lie on the same ground plane and estimate a homography $\mathbf{H}$. Given $\mathbf{H}$, we compute the projected world coordinates $(\hat E_i,\hat N_i)$ from the image coordinates $(u_i,v_i)$ as follows:
\begin{equation}
\begin{bmatrix}
\hat E_i \\ \hat N_i \\ 1
\end{bmatrix}
\sim
\underbrace{
\begin{bmatrix}
h_{11} & h_{12} & h_{13}\\
h_{21} & h_{22} & h_{23}\\
h_{31} & h_{32} & h_{33}
\end{bmatrix}
}_{\mathbf{H}}
\begin{bmatrix}
u_i \\ v_i \\ 1
\end{bmatrix},
\end{equation}
where $\sim$ denotes equality up to scale. Dividing the first two components of the matrix product by the third gives the projected coordinates in meters:
\begin{equation}
\begin{aligned}
\hat E_i &= \frac{h_{11}u_i+h_{12}v_i+h_{13}}{h_{31}u_i+h_{32}v_i+h_{33}},\\
\hat N_i &= \frac{h_{21}u_i+h_{22}v_i+h_{23}}{h_{31}u_i+h_{32}v_i+h_{33}}.
\end{aligned}
\label{eq:supp_image_to_world}
\end{equation}

The reprojection error for point pair $i$ is the Euclidean distance between the projected position $(\hat E_i,\hat N_i)$ and the reference position $(E_i,N_i)$. We use RANSAC in OpenCV's \texttt{findHomography} function to select point pairs with a reprojection error threshold of 0.5\,m.

Using these retained pairs, indexed by $\mathcal{I}$, we estimate the final homography $\mathbf{H}^{*}$ by minimizing the sum of squared reprojection errors
\begin{equation}
\mathbf{H}^{*} = \operatorname*{arg\,min}_{\mathbf{H}}
\sum_{i\in\mathcal{I}}
\left[(\hat E_i-E_i)^2+(\hat N_i-N_i)^2\right].
\label{eq:supp_homography_fit}
\end{equation}

\paragraph{Reprojection Error at Reference Points.}
We assess how closely the final homography $\mathbf{H}^{*}$ maps the retained image points to their Google Earth reference positions by measuring the reprojection error in meters. We project each image point $(u_i,v_i)$ to $(\hat E_i,\hat N_i)$ using $\mathbf{H}^{*}$ in \cref{eq:supp_image_to_world}. Its reprojection error is the Euclidean distance in meters between this projected position and the reference position $(E_i,N_i)$ obtained from \cref{eq:supp_local_world},
\begin{equation}
\varepsilon_i = \sqrt{(\hat E_i-E_i)^2 + (\hat N_i-N_i)^2}.
\label{eq:supp_reprojection_error}
\end{equation}
\Cref{tab:supp_homography_error} reports reprojection errors for the 56 pairs retained by RANSAC. For each scene, we compute the root mean squared error (RMSE) over its retained pairs $\mathcal{I}$ as
\begin{equation}
\mathrm{RMSE} = \sqrt{\frac{1}{|\mathcal{I}|} \sum_{i\in\mathcal{I}} \varepsilon_i^2},
\end{equation}
where $|\mathcal{I}|$ is the number of retained pairs in that scene. The scene-wise RMSE ranges from 0.086 to 0.183\,m. Pooling the squared errors over all 56 retained pairs gives an RMSE of 0.137\,m. These low errors indicate that the final homographies map the retained image points closely to their Google Earth reference positions and are accurate enough to reconstruct pedestrian trajectories in world coordinates within the annotated regions.

\paragraph{Pedestrian Positions to World Coordinates.}
We use the bottom center of each pedestrian's 2D bounding box to approximate their position on the ground. Let $(x_{\mathrm{tl}},y_{\mathrm{tl}})$ and $(x_{\mathrm{br}},y_{\mathrm{br}})$ denote the normalized coordinates of the top-left and bottom-right corners of the box. For an original image of width $W$ and height $H_{\mathrm{img}}$, the pixel coordinates of the bottom center are
\begin{equation}
(u,v)=\left(\frac{x_{\mathrm{tl}}+x_{\mathrm{br}}}{2}\,W,\;y_{\mathrm{br}}H_{\mathrm{img}}\right).
\end{equation}
We apply the scene-specific homography $\mathbf{H}^{*}$ to $(u,v)$ using \cref{eq:supp_image_to_world} to obtain the pedestrian's world coordinates in meters.

To limit extrapolation, we construct the smallest axis-aligned rectangle containing all reference points in the image. We expand it by $0.03W$ on the left and right and by $0.03H_{\mathrm{img}}$ on the top and bottom, clipping the rectangle to the image boundaries. We include a pedestrian observation in the exported trajectories only if its bottom-center point $(u,v)$ lies within this region.

\begin{table}[t]
    \centering
    \caption{\textbf{Homography reprojection errors.}
    Errors are measured in meters in world coordinates at the reference points.
    $N_{\mathrm{in}}$ denotes the number of annotated reference point pairs
    retained as inliers by RANSAC.}
    \label{tab:supp_homography_error}
    \small
    \setlength{\tabcolsep}{3pt}
    \renewcommand{\arraystretch}{1.08}
    \begin{tabular*}{\columnwidth}{@{\extracolsep{\fill}}cccccc@{}}
        \toprule
        Scene & $N_{\mathrm{in}}$ & Mean & Median & Max & RMSE \\
        \midrule
        Univ. Canteen  &  9 & 0.097 & 0.059 & 0.255 & 0.128 \\
        OCT Harbour    &  8 & 0.103 & 0.064 & 0.196 & 0.124 \\
        Xili Crossroad & 12 & 0.087 & 0.052 & 0.204 & 0.112 \\
        Primary School &  7 & 0.066 & 0.067 & 0.155 & 0.086 \\
        Univ. Campus   & 13 & 0.173 & 0.157 & 0.287 & 0.183 \\
        Huaqiangbei    &  7 & 0.120 & 0.100 & 0.267 & 0.146 \\
        \midrule
        All (pooled) & 56 & 0.112 & 0.098 & 0.287 & 0.137 \\
        \bottomrule
    \end{tabular*}
\end{table}

\begin{table}[t]
    \centering
    \caption{\textbf{Distribution of per-pedestrian mean speeds (m/s).}
    $N$ denotes the number of pedestrians with at least 10 valid steps
    between consecutive frames.}
    \label{tab:supp_speed}
    \small
    \setlength{\tabcolsep}{5pt}
    \renewcommand{\arraystretch}{1.08}
    \begin{tabular*}{\columnwidth}{@{\extracolsep{\fill}}ccccc@{}}
        \toprule
        Scene & $N$ & P25 & Median & P75 \\
        \midrule
        Univ. Canteen  &   176 & 1.18 & 1.38 & 1.65 \\
        OCT Harbour    &   545 & 0.37 & 0.75 & 1.03 \\
        Xili Crossroad &   442 & 0.86 & 1.07 & 1.44 \\
        Primary School &   369 & 0.12 & 0.43 & 0.85 \\
        Univ. Campus   &   321 & 0.45 & 0.92 & 1.24 \\
        Huaqiangbei    & 1,610 & 0.87 & 1.26 & 1.59 \\
        \bottomrule
    \end{tabular*}
\end{table}

\begin{figure*}[!t]
    \centering
    \includegraphics[width=\textwidth]{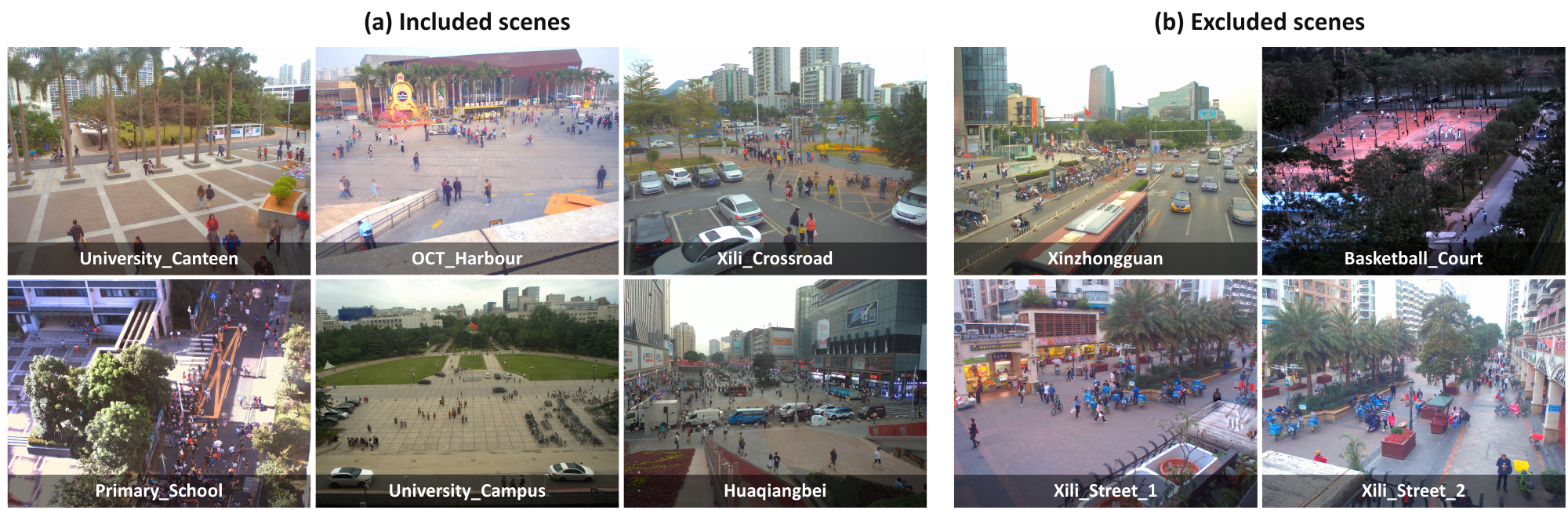}
    \caption{\textbf{Scenes included in CAT.} Representative frames from the six included scenes (a) and the four excluded scenes (b).}
    \label{fig:cat_scenes}
\end{figure*}

\paragraph{Speed Plausibility Check.}
We examine pedestrian speeds as an additional check on the spatial scale of the reconstructed trajectories. For a given pedestrian, let $\mathbf{x}_k=(E_k,N_k)$ denote their world coordinates in meters at frame $k$. When the pedestrian is observed in two consecutive frames, we compute their speed $s_k$ from the displacement between the two positions:
\begin{equation}
s_k=\frac{\lVert\mathbf{x}_{k+1}-\mathbf{x}_k\rVert_2}{\Delta t_{\mathrm{data}}},
\end{equation}
where $\Delta t_{\mathrm{data}}=0.5$\,s at the dataset sampling rate of 2\,FPS. We include pedestrians with at least 10 such frame pairs and compute each pedestrian's mean speed over all valid pairs. This calculation includes all mobility modes and stationary periods. \Cref{tab:supp_speed} summarizes the distribution of these mean speeds within each scene.

The median of these mean speeds ranges from 0.92 to 1.38\,m/s across \textit{University Canteen}, \textit{Xili Crossroad}, \textit{University Campus}, and \textit{Huaqiangbei}. The corresponding medians are lower in \textit{OCT Harbour} (0.75\,m/s) and \textit{Primary School} (0.43\,m/s), consistent with slower movement or more time spent stationary. Together, the low reprojection errors at the retained reference points and the plausible pedestrian speeds across scenes support the use of the estimated homographies to construct pedestrian trajectories in world coordinates.

\subsection{CAT Dataset Construction}
\label{sec:supp_cat_construction}

We describe how we select six PANDA scenes for our dataset, assign initial pedestrian attribute labels, and review these labels through manual annotation. We then explain how we construct the scene semantic maps and summarize the resulting dataset with scene-level statistics.

\paragraph{Data Sources and Scene Selection.}
We construct CAT using PANDA video frames and tracking annotations obtained through the GigaMOT Challenge on the GigaVision platform. We inspect each pedestrian across multiple frames to identify their attributes and use the tracking annotations to align these attributes to the corresponding trajectory. The PANDA videos were recorded in public areas where photography was officially approved, and faces are already blurred in the frames we obtained through the challenge.

The training split includes pedestrian IDs, frame-by-frame bounding boxes, face orientations, and occlusion levels. The test split contains images and basic sequence information but does not provide ground-truth tracking annotations. We therefore consider the 10 training scenes for dataset construction.

As shown in \cref{fig:cat_scenes}, we select six scenes that capture everyday pedestrian movement and contain ground reference points suitable for homography estimation. We exclude \textit{Xinzhongguan} because changes in ground elevation make it difficult to represent the walkable area with a single homography. We exclude \textit{Basketball Court} because it primarily captures sports activities. For \textit{Xili Street 1} and \textit{Xili Street 2}, we could not identify suitable ground points that could be matched between the video frames and satellite imagery.

\suppressfloats[t]
\begin{figure}[t]
    \centering
    \includegraphics[width=\columnwidth]{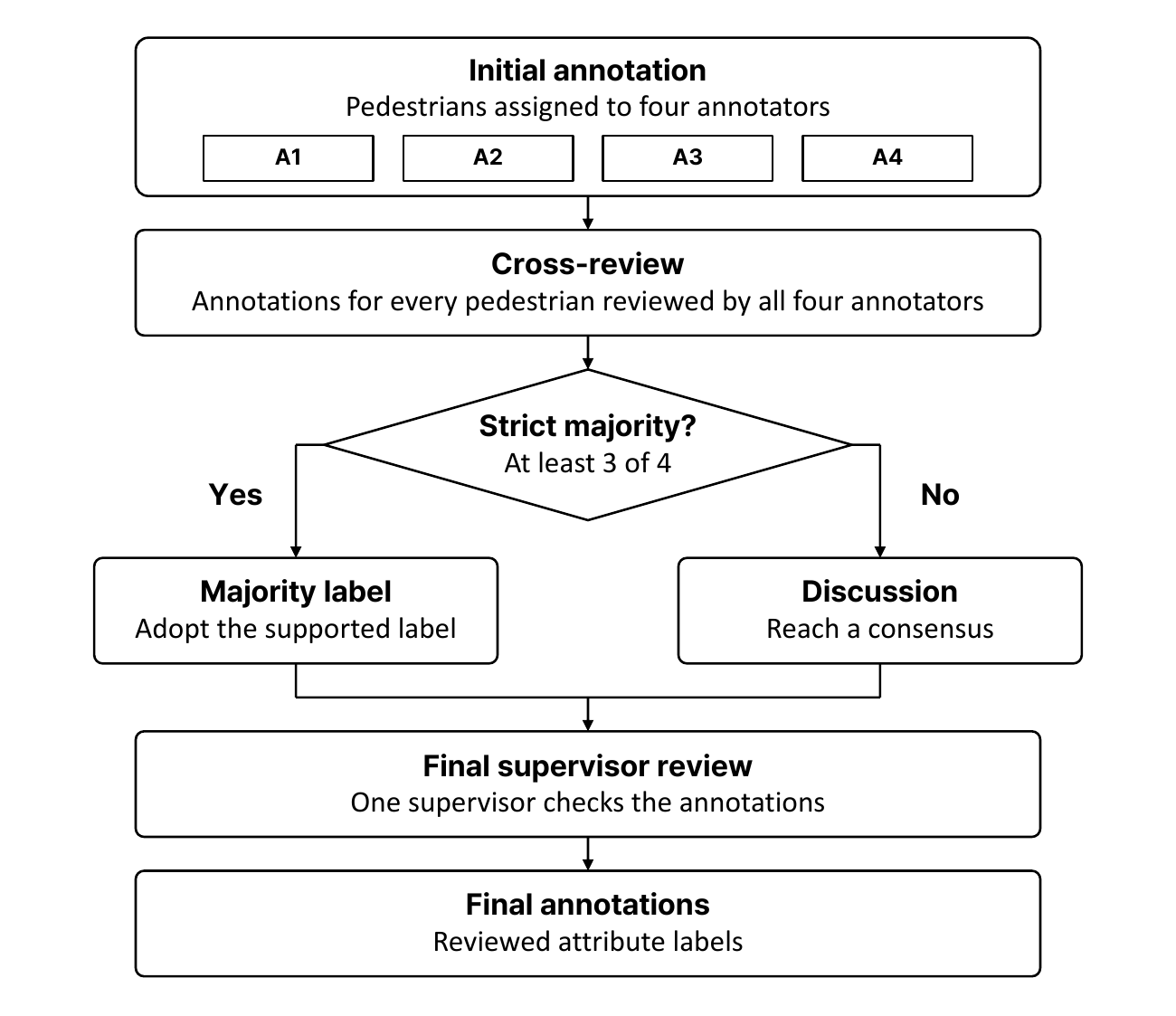}
    \caption{\textbf{Annotation and review workflow.}}
    \label{fig:annotation_workflow}

    \vspace{1.2em}
    \includegraphics[width=\columnwidth]{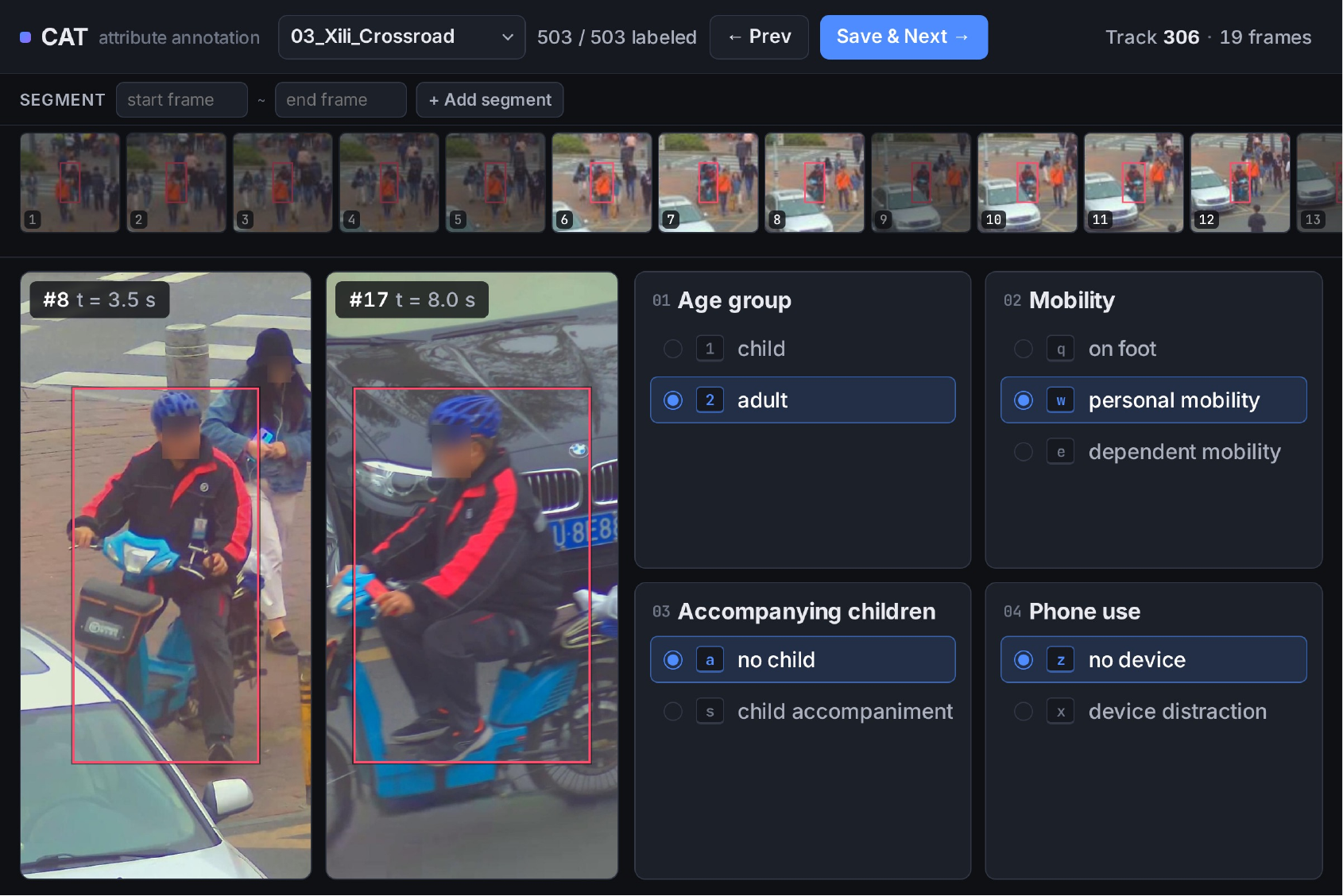}
    \caption{\textbf{Web-based annotation interface.}}
    \label{fig:annotation_interface}
\end{figure}
\begin{figure}[t]
    \centering
    \includegraphics[width=\columnwidth]{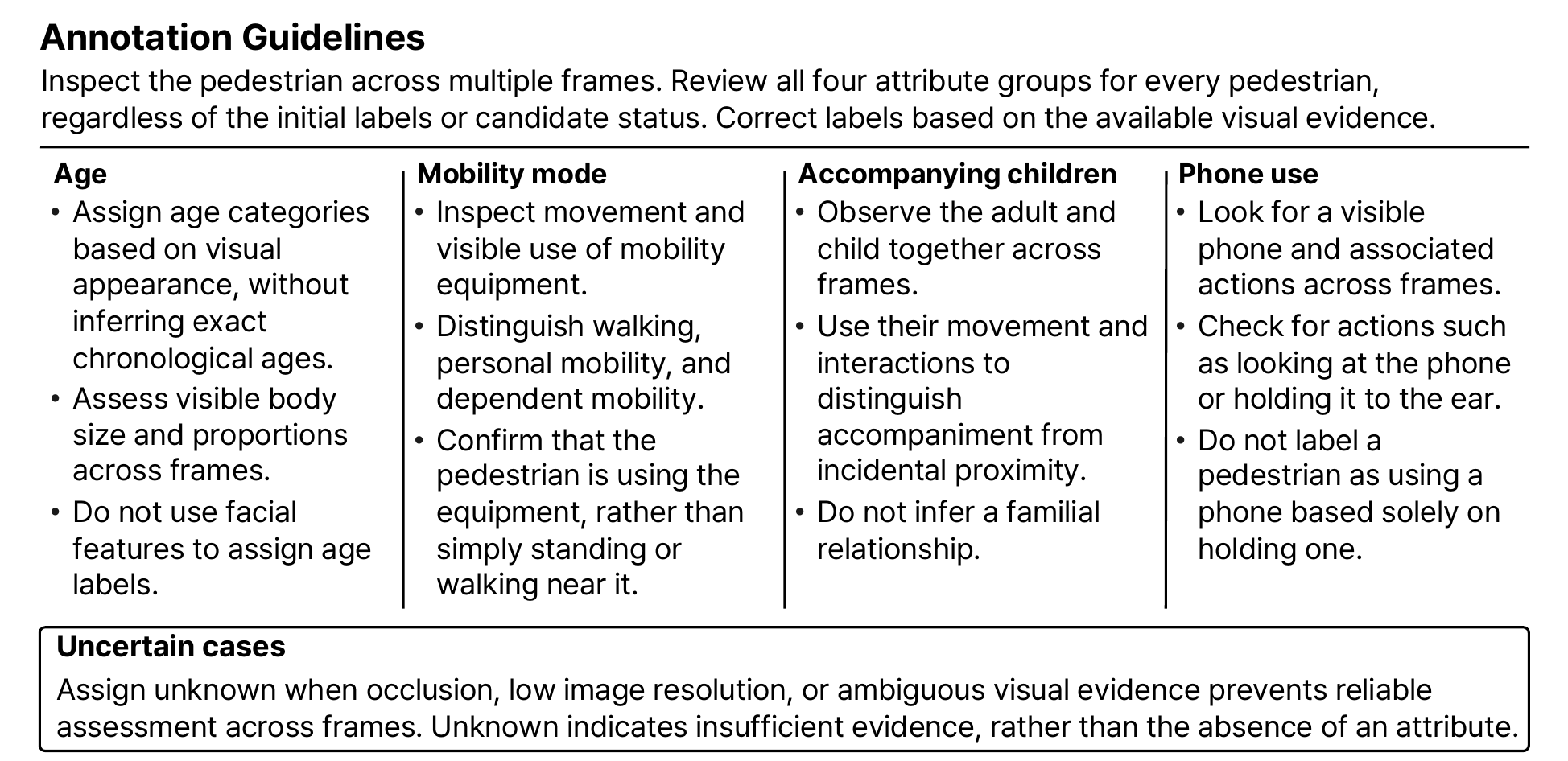}
    \caption{\textbf{Manual annotation guidelines for CAT.}}
    \label{fig:annotation_guidelines}

    \vspace{1.2em}
    \includegraphics[width=\columnwidth]{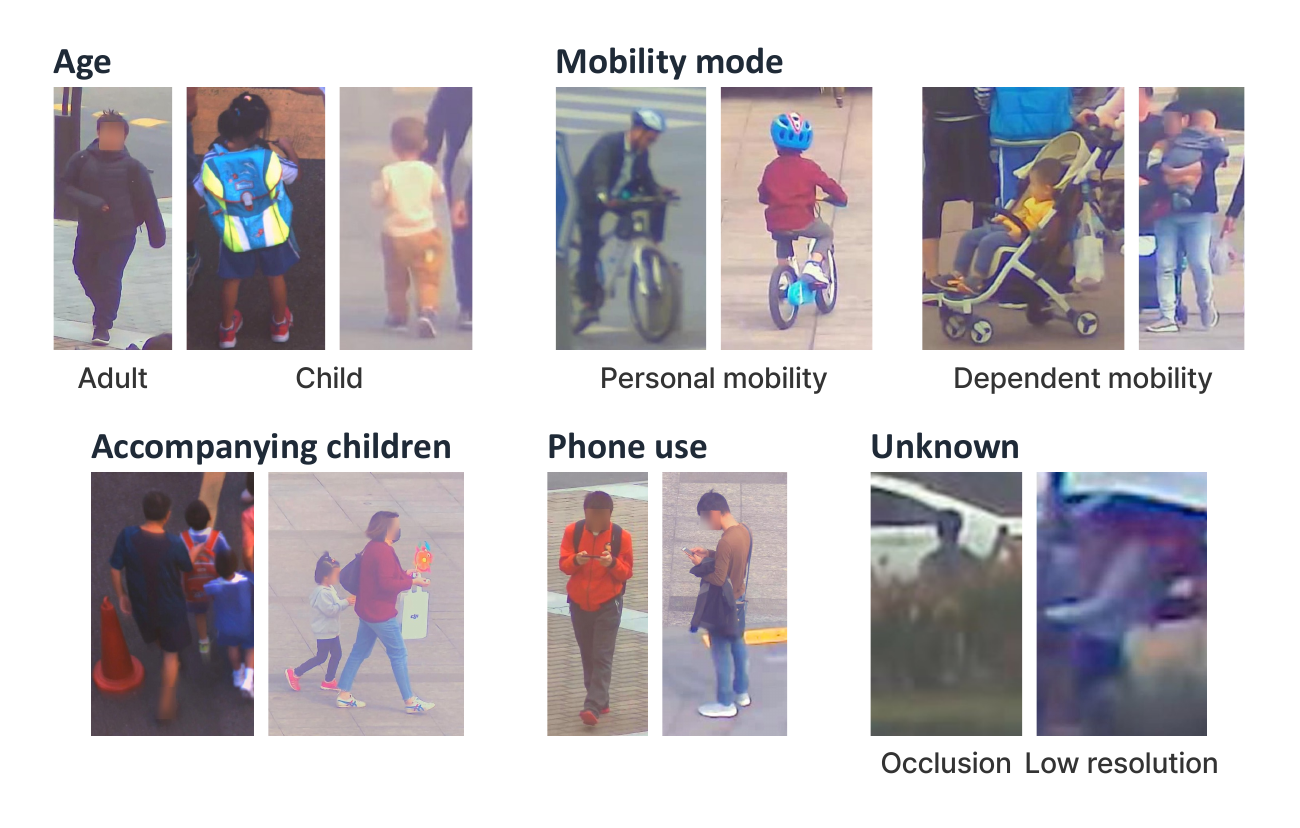}
    \caption{\textbf{Examples of pedestrian attribute annotations in CAT.}}
    \label{fig:annotation_examples}
\end{figure}

The six selected scenes cover university areas, commercial spaces, a primary school, and an urban intersection. \textit{University Canteen} and \textit{University Campus} capture pedestrian movement around dining areas and along campus walkways, respectively. \textit{OCT Harbour} features an open commercial plaza, while \textit{Huaqiangbei} captures pedestrian traffic along shopping streets. \textit{Primary School} includes children and accompanying adults near the school entrance, and \textit{Xili Crossroad} captures pedestrians crossing an intersection in several directions. These settings allow us to examine how pedestrian composition varies across scenes and how movement varies with pedestrian attributes.

\begin{figure*}[!t]
    \centering
    \includegraphics[width=\textwidth]{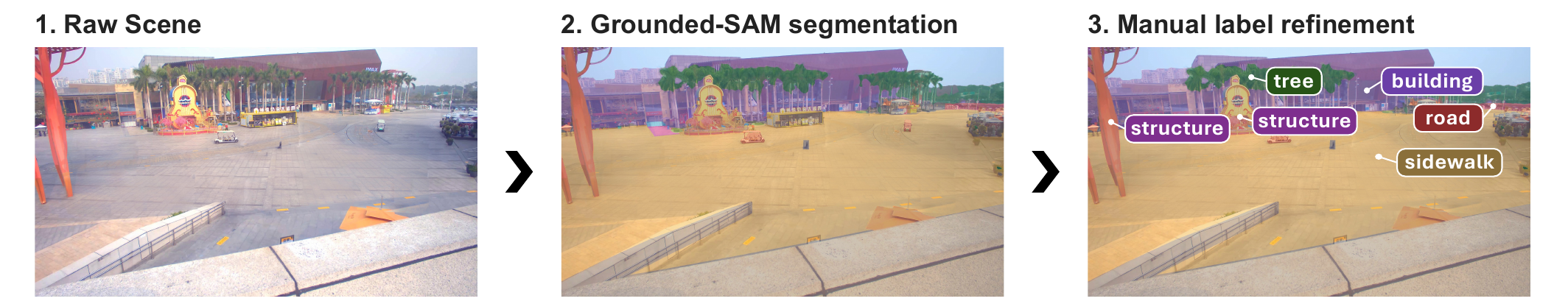}
    \caption{\textbf{Scene semantic annotation process.}}
    \label{fig:semantic_map_construction}
\end{figure*}

\begin{figure}[!t]
    \centering
    \includegraphics[width=\columnwidth]{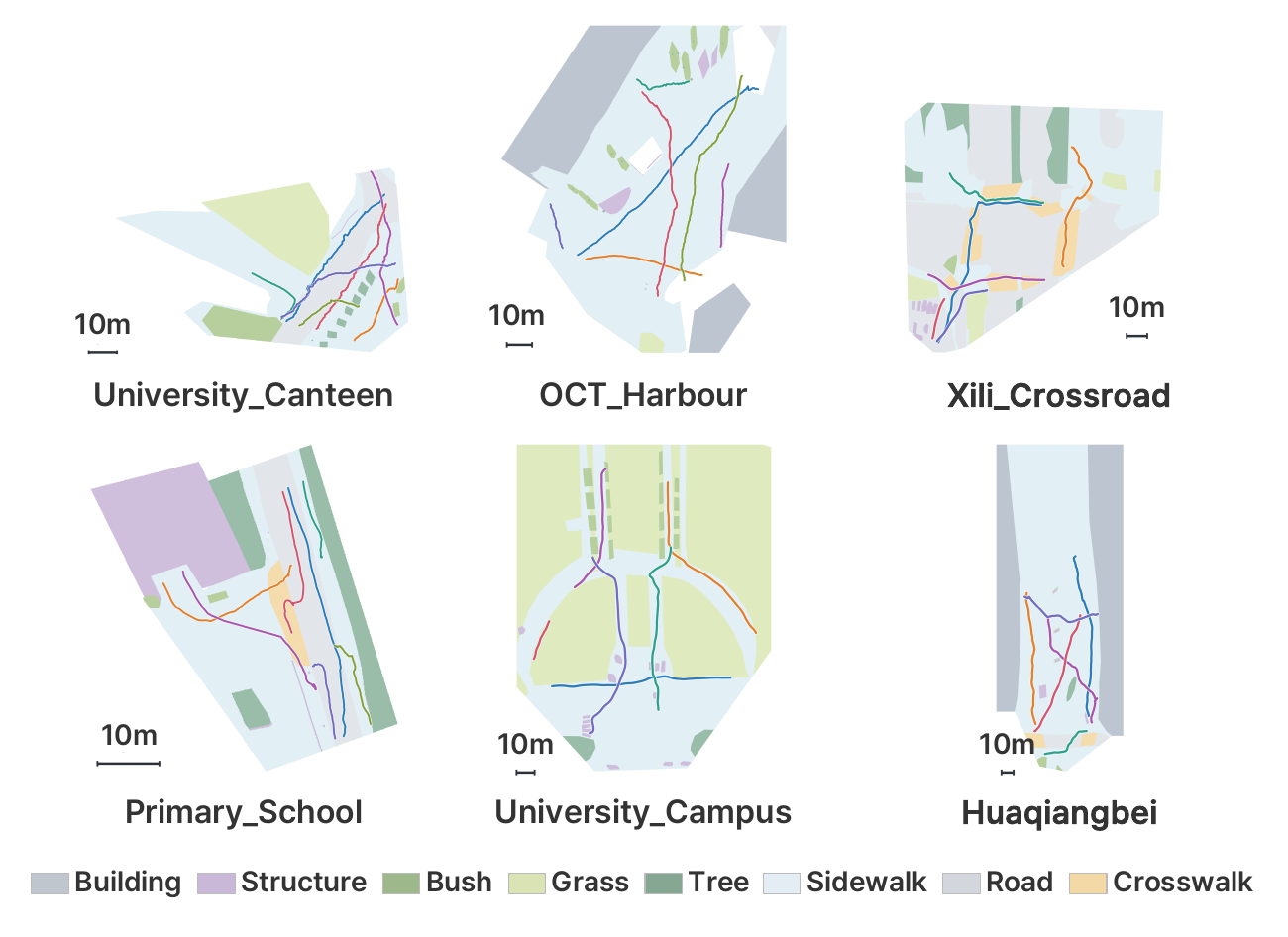}
    \caption{\textbf{Scene semantic maps of CAT.}
    Region colors indicate semantic classes.}
    \label{fig:scene_semantic_maps}
\end{figure}

\paragraph{Candidate Labeling Before Manual Review.}
Before manual annotation, we use a model and predefined rules for each attribute group to assign initial labels to each pedestrian across all four groups.

(1) Age: We identify child candidates using the ratio of head height to full-body height, following prior work on head-to-body proportions~\cite{ince2014child}. We mark pedestrians with a ratio of $0.20$ or higher as child candidates.

(2) Mobility mode: We apply YOLOE~\cite{wang2025yoloe} to an enlarged crop around each pedestrian, using text prompts for bicycles, motorcycles and strollers. We mark a pedestrian as a personal-mobility candidate if a detected bicycle or motorcycle overlaps the lower part of their bounding box across multiple frames. We apply the same overlap criterion to strollers to identify dependent-mobility candidates. To identify pedestrians who may be carried, we also mark those whose bounding boxes remain largely within another pedestrian's bounding box across multiple frames as dependent-mobility candidates.

(3) Accompanying children: We compare the world-coordinate trajectories of child candidates with those of adults in the same scene. We use a distance threshold of 1.2\,m, corresponding to the approximate outer boundary of personal distance in Hall's proxemic zones. We mark an adult as potentially accompanying a child if the adult is within this distance of a child candidate in at least 10 frames. At 2\,FPS, this represents 5\,s of observations in total. We allow these observations to occur in separate intervals to capture repeated proximity while reducing candidates arising from brief encounters.

(4) Phone use: We identify phone-use candidates using PANDA's face orientation annotations, which represent horizontal facing directions in eight discrete bins. We estimate each pedestrian's walking direction from changes in their world coordinates across frames. We mark a pedestrian when their face orientation stays within $45^\circ$ of their walking direction and changes infrequently across multiple frames.

These candidate labels provide a starting point for manual annotation. In the next stage, annotators review all four attributes for every pedestrian, including those with no marked attributes, and assign the final labels based on visual evidence across frames.

\paragraph{Annotation Procedure and Review.}
We hired four annotators with academic backgrounds in the social sciences to label the pedestrian attributes discussed in the introduction. \cref{fig:annotation_workflow} summarizes the annotation and review process. Each annotator is initially responsible for a subset of pedestrians. To reduce annotation errors, we designed the web-based interface shown in \cref{fig:annotation_interface} to display an enlarged view of the selected pedestrian while preserving the original image resolution. Annotators can inspect every frame in which the pedestrian appears, allowing them to follow the pedestrian's movement and interactions over time. They use this interface to review and, where necessary, correct the initial labels for all four attribute groups according to the guidelines in \cref{fig:annotation_guidelines}.

\begin{table}[!t]
    \centering
    \input{tables/scene_statistics}
\end{table}

The annotators then cross-check one another's work so that all four review the labels for every pedestrian. For each attribute, we retain a label if at least three annotators agree. If no label receives three votes, the annotators review the visual evidence together and discuss the case. After discussion, the annotators assign the label they agree on or mark the attribute as \textit{unknown} if occlusion, low image resolution, or ambiguous visual evidence prevents them from reaching a consensus. Once the annotators have assigned a label to each attribute, a supervisor reviews the completed annotations. \cref{fig:annotation_examples} shows representative annotations, including examples labeled as \textit{unknown}.

\paragraph{Scene Semantic Map Construction.}
For each scene, we use Grounded-SAM to generate initial segmentation masks for the eight semantic classes in a single video frame. We review the masks and their class labels against the original frame, correcting region boundaries and labels where needed. \cref{fig:semantic_map_construction} illustrates this process from the original frame to the manually refined annotation. We then project the labeled regions from image coordinates to world coordinates using the same scene-specific homography used to reconstruct the pedestrian trajectories. This allows us to overlay pedestrian trajectories on the semantic map and see which regions pedestrians move through. \cref{fig:scene_semantic_maps} shows the resulting bird's-eye-view maps, with region colors indicating semantic classes and lines showing pedestrian trajectories.

\paragraph{The CAT Dataset Summary.}
\label{sec:CAT_scene}
The final dataset contains 378,400 observations of 4,048 pedestrians across six scenes. \Cref{tab:scene_statistics} summarizes pedestrian counts, trajectory durations, and speeds for each scene. Across all tracks, the median trajectory duration is 37.00\,s. For speed, we use the same tracks as in \cref{tab:supp_speed}, each with at least 10 valid steps between consecutive frames. The median of the per-pedestrian mean speeds across these tracks is 1.05\,m/s. Scene-level medians range from 23.00 to 54.50\,s for trajectory duration and from 0.43 to 1.38\,m/s for speed. The variation in speed across scenes is broadly consistent with field observations of pedestrian movement in urban settings~\cite{willis2004human}. Each trajectory is linked to manually reviewed labels for all four attribute groups and shares a world coordinate system with the scene's semantic map. Together, these data allow us to examine how pedestrian composition varies across scenes and how movement varies with pedestrian attributes and surroundings.

\input{tables/scene_attr_statistics}

\begin{figure}[!t]
    \centering
    \includegraphics[
        width=\columnwidth,
        height=0.88\textheight,
        keepaspectratio
    ]{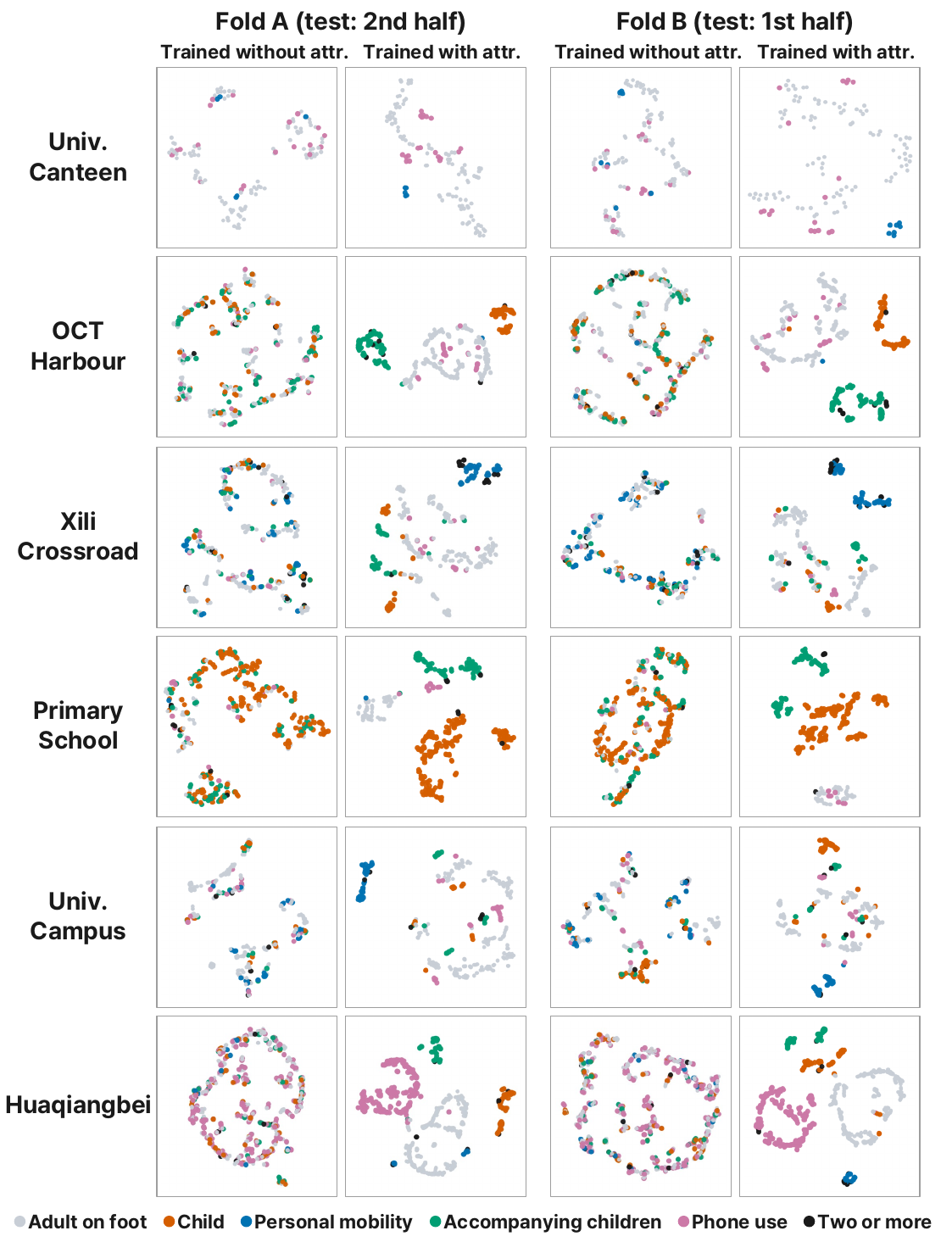}
    \caption{\textbf{Scene-wise t-SNE visualization of movement features.}
    Each row shows one CAT scene. For each temporal fold, we compare features
    from models trained with and without pedestrian attributes.
    Colors indicate pedestrian categories.}
    \label{fig:attr_tsne}
\end{figure}

\subsection{Scene-wise Attribute Analysis}
\label{sec:supp_scene_attr}

We examine whether the attribute-dependent movement patterns observed in the pooled CAT statistics also appear within individual scenes. \Cref{tab:scene_attr_stats} shows behavioral statistics for each pedestrian attribute group across the six scenes. Each group's share is expressed as a percentage of all pedestrians in the scene.

We compute behavioral measures for pedestrians with at least 20 observations. For each pedestrian, we estimate speed from displacements over 1\,s and take the median over moving intervals. During movement, we compute acceleration as the absolute difference between two such speed estimates taken 1\,s apart, divided by 1\,s. For both speed and acceleration, we report the mean and standard deviation of the per-pedestrian values within each group. We also compute each pedestrian's median distance to the nearest other pedestrian across frames and the median duration of their continuous stops. For these two measures, we report the group median together with the first and third quartiles. The companion rate is the percentage of a pedestrian's moving frames in which another pedestrian moves with them within 1.5\,m. We report the group mean of these percentages. Behavioral statistics based on fewer than 10 pedestrians are marked as ``--''.

The main behavioral patterns observed in the pooled data also appear within individual scenes. Children move more slowly on average than adults on foot in four of the five scenes with reported speeds. In \textit{OCT Harbour}, the two groups have nearly identical mean speeds (0.86 vs.\ 0.85\,m/s), which may reflect a shared tendency to walk slowly in the park. Pedestrians accompanying children also show movement patterns similar to those of children. In \textit{Primary School}, for example, the two groups have similar median neighbor distances (0.44 and 0.43\,m, respectively) and high companion rates (87\% and 79\%), compared with 0.73\,m and 41\% for adults on foot. Personal mobility users move faster on average and have greater median neighbor distances than adults on foot in all three scenes with reported values. Phone users have greater median neighbor distances than adults on foot in all six scenes and longer median stop times in five. Both groups have their longest median stop times in \textit{Primary School} (51.4\,s for phone users and 36.8\,s for adults on foot), consistent with waiting near the school entrance. The scene-wise t-SNE visualizations in \cref{fig:attr_tsne} complement these statistics by illustrating how movement patterns vary across pedestrian attribute groups within each scene.

\subsection{Limitations}
To capture how pedestrian movement varies across diverse settings and pedestrian attributes, we select six scenes spanning university areas, commercial spaces, a primary school, and an urban intersection. We use findings from social science research and empirical studies of pedestrian behavior to select four attribute groups related to movement. To improve annotation reliability, four annotators with academic backgrounds in the social sciences review each pedestrian’s labels across multiple frames and cross-check one another’s assessments. 
Despite these efforts, the six scenes represent a limited range of populations and environments, and the four attribute groups capture only some aspects of pedestrian interaction. For example, CAT records whether an adult accompanies a child, but it lacks interaction annotations of gestures, conversations, or physical contact. This makes it difficult to examine how pedestrians change their speed or direction during specific interactions. Future work could extend CAT to more diverse populations and settings and record who interacts with whom, how, and when.

\section{Method Details}
\label{app:architecture}

\subsection{Notation and Coordinates}
\label{app:notation}
We use the notation of \cref{sec:method}. In particular, $a^i$ denotes the concatenation of one-hot vectors specifying agent $i$'s category within each of the four attribute groups, whereas $e^i=W_a a^i$ denotes its learned embedding. The embedding conditions goal prediction, social interaction, and displacement prediction.

The origin $o^i$, current position $x_t^i$, and current goal $g_t^i$ are defined in world coordinates. The initial goal sampled during emission is $g^i$. For an agent entering at time $t_0^i$, we have $g_{t_0^i}^i=g^i$. Its goal may subsequently change, but $a^i$ and $e^i$ remain fixed throughout its lifetime. The vector $v_t^i$ denotes velocity, and $\Delta_t^i$ denotes the predicted displacement over one motion update. We use $\Delta t$ for the duration of that update.

Let $\Omega$ denote the set of grid cells in the scene map, indexed by $u$ and $v$ as in \cref{eq:origin_distribution}. For a world-coordinate position $x$, $\psi(x)$ gives its grid coordinates, and $p(x)$ gives the corresponding normalized coordinates. We write $p_t^i=p(x_t^i)$ and compute the relative position in \cref{eq:social_bias} as
\begin{equation}
r_t^{ij}=p_t^j-p_t^i.
\label{eq:app_relative_position}
\end{equation}
Social attention uses normalized coordinates, while trajectories and displacements use world coordinates. Spatial features at a world-coordinate position are obtained from the corresponding location in the feature maps.

\begin{table*}[t]
\centering
\small
\caption{Notation used in the main paper and this appendix.}
\label{tab:app_notation}
\begin{tabularx}{\textwidth}{@{}lX@{}}
\toprule
Symbol & Meaning \\
\midrule
$S,M$ & Semantic map and traversable-region mask. \\
$H_{\mathrm{ego}},H_{\mathrm{nbr}}$ & Heatmaps of recent target-agent and neighbor positions. \\
$F_{\mathrm{global}},F_{\mathrm{spatial}}$ & Global scene features and multi-scale spatial features from $\theta_{\mathrm{enc}}$. \\
$\Omega,\psi(x),p(x)$ & Set of grid cells, grid coordinates of $x$, and corresponding normalized coordinates. \\
$o^i,g^i$ & Sampled origin and initial goal of agent $i$. \\
$a^{i,(r)},a^i$ & One-hot category vector for attribute group $r$ and concatenated vector across all four groups. \\
$e^i=W_a a^i$ & Learned attribute embedding. \\
$x_t^i,g_t^i,v_t^i$ & Current position, goal, and velocity in world coordinates. \\
$r_t^{ij},B_t^{ij}$ & Normalized relative position and attention bias for query agent $i$ attending to agent $j$. \\
$X_t$ & Agent-token matrix, with one row per active agent. \\
$h_t^i,z_t^i$ & Social feature and concatenated input to the motion head. \\
$\Delta_t^i,\Delta t$ & Predicted one-step displacement and duration of a motion update. \\
$\mathcal{X}^i,\widehat{\mathcal{X}}^i$ & Observed and predicted trajectory sequences used in reconstruction losses. \\
\bottomrule
\end{tabularx}
\end{table*}

\subsection{Shared Scene Encoder}
\label{app:shared_features}

The U-Net encoder $\theta_{\mathrm{enc}}$ takes the channel-wise concatenation $[M,S,H_{\mathrm{ego}},H_{\mathrm{nbr}}]$, as in \cref{eq:shared_encoder}. The global features $F_{\mathrm{global}}$ capture scene context, while the multi-scale spatial features $F_{\mathrm{spatial}}$ preserve local semantic and spatial information.

During emission, position histories are unavailable for newly arriving agents. We therefore set both position heatmaps to zero and compute the scene features as
\begin{equation}
\begin{aligned}
&(F_{\mathrm{global}},F_{\mathrm{spatial}})\\
&\qquad=\theta_{\mathrm{enc}}([M,S,0,0]).
\end{aligned}
\label{eq:app_emission_features}
\end{equation}
These features are used to predict origin and stationary occupancy maps, attribute distributions, and goal heatmaps.

During motion generation, the heatmaps encode the recent positions of the target agent and its neighbors. The resulting features therefore depend on the target agent and time step, although these indices are omitted for simplicity. The same encoder parameters are used for emission and motion generation.

\subsection{Origin Maps and Attribute Representation}
\label{app:origin_attributes}

The origin head $\theta_{\mathrm{origin}}$ jointly predicts an all-origin logit map $Z_{\mathrm{all}}$, attribute-specific logit maps $Z_{\mathrm{attr}}$, and a stationary occupancy logit map $Z_{\mathrm{occ}}$ from the emission features. The attribute-specific maps provide auxiliary supervision, while the stationary occupancy map describes locations where pedestrians tend to stop.

The output channels are jointly supervised using weighted binary cross-entropy with logits. The loss is averaged over all channels and grid cells, so stationary occupancy supervision is included in the spawn loss. For origin sampling, we use the normalized distribution in \cref{eq:origin_distribution}.

For each of the four attribute groups $r\in\{1,\ldots,4\}$, let $C_r$ denote the number of categories and $a^{i,(r)}$ the one-hot vector specifying agent $i$'s category. We concatenate these vectors to form $a^i$ and project it to the embedding $e^i$:
\begin{equation}
\begin{aligned}
a^{i,(r)}&\in\{0,1\}^{C_r},\qquad
\sum_{k=1}^{C_r}a_k^{i,(r)}=1,\\
a^i&=\operatorname{concat}_{r=1}^{4}a^{i,(r)},\\
e^i&=W_a a^i.
\end{aligned}
\label{eq:app_attribute_code}
\end{equation}
The concatenated vector has dimension $C_a=\sum_{r=1}^{4}C_r$. For an embedding dimension $d_e$, the projection matrix is $W_a\in\mathbb{R}^{d_e\times C_a}$.

The four groups represent age, mobility mode, accompanying children, and phone use. Each agent has one category within each group, while attributes from different groups may co-occur. The vector $a^i$ specifies these categories, and $e^i$ provides the learned representation used to condition goal and motion generation.

\subsection{Scene Cross-Attention}
\label{app:scene_attention}

For origin $o^i$, we concatenate the local feature $F_{\mathrm{spatial}}(o^i)$ with the coordinates of $o^i$ to form the row vector $q_{\mathrm{in}}^i$. We flatten the spatial dimensions of $F_{\mathrm{global}}$ to obtain the scene-token matrix $G$, with one token per row. For attention head $m$, the projections are
\begin{equation}
\begin{aligned}
Q_i^{\mathrm{sc},m}
&=q_{\mathrm{in}}^i W_Q^{\mathrm{sc},m},\\
K^{\mathrm{sc},m}
&=G W_K^{\mathrm{sc},m},\\
V^{\mathrm{sc},m}
&=G W_V^{\mathrm{sc},m}.
\end{aligned}
\label{eq:app_scene_projections}
\end{equation}
The query depends on the origin, whereas the keys and values describe the scene. Each head computes
\begin{equation}
O_i^{\mathrm{sc},m}
=\operatorname{softmax}\!\left(
\frac{Q_i^{\mathrm{sc},m}(K^{\mathrm{sc},m})^\top}
{\sqrt{d_{\mathrm{sc}}}}
\right)V^{\mathrm{sc},m},
\label{eq:app_scene_attention}
\end{equation}
where the softmax is taken over scene tokens and $d_{\mathrm{sc}}$ is the query/key dimension of each head, as in \cref{eq:attribute_distribution}.

We concatenate the head outputs and apply an output projection to obtain an origin-conditioned feature $c_{\mathrm{sc}}^i$. The classifier for attribute group $r$ then predicts
\begin{equation}
\pi^{i,(r)}
=\operatorname{softmax}(f_r(c_{\mathrm{sc}}^i)),
\label{eq:app_attribute_probabilities}
\end{equation}
where $f_r$ produces $C_r$ category logits.

During training, the classifiers are supervised at annotated origins using the corresponding attribute labels. During generation, we sample one category from each $\pi^{i,(r)}$. The corresponding one-hot vectors form $a^i$, which is mapped to $e^i$ by $W_a$. Thus, $p_a$ in \cref{eq:emission} defines a distribution over categorical attribute vectors, and the embedding is a deterministic function of the sampled attributes. The auxiliary maps $Z_{\mathrm{attr}}$ receive binary cross-entropy supervision, while the four classifiers receive categorical cross-entropy supervision.

\subsection{Attribute-Conditioned Goal Prediction}
\label{app:goal_prediction}

The goal head uses the emission features computed with zero position heatmaps. Following \cref{eq:film}, we condition the global scene features on the attribute embedding $e^i$ using FiLM:
\begin{equation}
\widetilde{F}^{\,i}
=\gamma(e^i)\odot F_{\mathrm{global}}
+\beta(e^i).
\label{eq:app_goal_film}
\end{equation}
The functions $\gamma$ and $\beta$ produce channel-wise scales and shifts, respectively. These values are broadcast over spatial locations.

The goal head maps $\widetilde{F}^{\,i}$ to goal logits $Z_{\mathrm{goal}}^i$. Applying the sigmoid function gives the predicted heatmap:
\begin{equation}
P_{\mathrm{goal}}^i(u)
=\sigma\!\left(Z_{\mathrm{goal}}^i(u)\right).
\label{eq:app_goal_heatmap}
\end{equation}
The heatmap is supervised using binary cross-entropy with target heatmaps constructed from annotated goal locations. The complete goal-training objective is detailed in \cref{eq:app_goal_loss}.

The predicted heatmap depends on the scene features and attribute embedding. The origin or current position is used during goal sampling to determine the distance constraint, rather than being supplied directly to the FiLM goal decoder.

During generation, we use the logits underlying the predicted heatmap to construct the goal-sampling distribution. Traversability penalties, a distance prior, and stationary occupancy scores are incorporated before normalization. The complete procedure is described in \cref{app:goal_sampling}.

\subsection{Attribute-Aware Social Attention}
\label{app:social_attention}

At time $t$, we represent agent $i$ using its normalized position $p_t^i$ and attribute embedding $e^i$:
\begin{equation}
(X_t)_{i,:}=f_{\mathrm{pos}}(p_t^i)+f_{\mathrm{attr}}(e^i).
\label{eq:app_agent_token}
\end{equation}
The two mappings have the same output dimension. Stacking the tokens as rows forms $X_t$, consistent with \cref{eq:social_attention}. For attention head $m$, the projections are
\begin{equation}
\begin{aligned}
Q_t^{\mathrm{soc},m}
&=X_tW_Q^{\mathrm{soc},m},\\
K_t^{\mathrm{soc},m}
&=X_tW_K^{\mathrm{soc},m},\\
V_t^{\mathrm{soc},m}
&=X_tW_V^{\mathrm{soc},m}.
\end{aligned}
\label{eq:app_social_projections}
\end{equation}
When referring to individual rows, the second subscript indexes the agent. The rows of $V_t^{\mathrm{soc},m}$ are attention value vectors, while $v_t^i$ denotes pedestrian velocity.

The head-specific bias and attention score for the ordered pair $(i,j)$ are
\begin{equation}
\begin{aligned}
B_t^{ij,m}
&=\phi_{\mathrm{rel}}^m(r_t^{ij})
+\phi_{\mathrm{pair}}^m(e^i,e^j,r_t^{ij}),\\
\ell_t^{ij,m}
&=\frac{
Q_{t,i}^{\mathrm{soc},m}
(K_{t,j}^{\mathrm{soc},m})^\top
}{\sqrt{d_{\mathrm{soc}}}}
+B_t^{ij,m}.
\end{aligned}
\label{eq:app_headwise_bias}
\end{equation}
Here, $d_{\mathrm{soc}}$ is the query/key dimension of each head. Each bias function supplies a scalar for each attention head, and the superscript $m$ indexes the corresponding output. The relative-position branch captures spatial dependence. The pair branch additionally allows the bias to vary with the attributes of both agents, even when their relative positions are identical.

Let $\mathcal{J}_t^i$ denote the agent indices admitted by the attention mask for query agent $i$. For $j\in\mathcal{J}_t^i$, we normalize the scores and aggregate the value vectors:
\begin{equation}
\begin{aligned}
\alpha_t^{ij,m}
&=\frac{\exp(\ell_t^{ij,m})}
{\sum_{k\in\mathcal{J}_t^i}\exp(\ell_t^{ik,m})},\\
O_{t,i}^{\mathrm{soc},m}
&=\sum_{j\in\mathcal{J}_t^i}
\alpha_t^{ij,m}V_{t,j}^{\mathrm{soc},m},\\
c_t^i
&=f_O\!\left(
\operatorname{concat}_m O_{t,i}^{\mathrm{soc},m}
\right).
\end{aligned}
\label{eq:app_headwise_attention}
\end{equation}
Here, $f_O$ denotes the output projection applied to the concatenated head outputs. These equations expand \cref{eq:social_attention} for an individual query agent, with the attention-head index made explicit. During generation, attention considers active agents without a distance cutoff, subject to the attention mask.

A GRU encodes the most recent $L$ velocity vectors. We combine its output with the social context using an MLP:
\begin{equation}
h_t^i=f_{\mathrm{soc}}\!\left(
\left[
c_t^i,
\operatorname{GRU}(v_{t-L+1:t}^i)
\right]
\right).
\label{eq:app_social_feature}
\end{equation}
The resulting $h_t^i$ represents both the target agent's recent motion and its interaction context.

\subsection{Displacement Prediction}
\label{app:displacement}

We form the motion input by concatenating the local spatial feature, social feature, current goal, and current velocity:
\begin{equation}
z_t^i=
[F_{\mathrm{spatial}}(x_t^i),h_t^i,g_t^i,v_t^i].
\label{eq:app_motion_input}
\end{equation}
The shared encoder takes the traversable-region mask $M$, the semantic map $S$, and heatmaps of the recent positions of the target agent and its neighbors. We sample the resulting spatial feature at the agent's current position $x_t^i$. The social feature is given by \cref{eq:app_social_feature}, and $g_t^i$ is the current goal in world coordinates.

The motion head $\phi_{\mathrm{mot}}$ conditions the concatenated features on $e^i$ using FiLM, as in \cref{eq:film}. The next-step displacement is
\begin{equation}
\Delta_t^i
=\exp(w^\top e^i)\,
\phi_{\mathrm{mot}}(z_t^i,e^i).
\label{eq:app_motion_prediction}
\end{equation}
The vector $w$ is learnable. The positive multiplier $\exp(w^\top e^i)$ adjusts the magnitude of the motion-head output while preserving its direction. Both the goal head and the motion head use FiLM conditioned on the attribute embedding $e^i=W_a a^i$.

The resulting $\Delta_t^i$ is the predicted one-step displacement in world coordinates. During generation, we add temporally correlated noise to $\Delta_t^i$ and adjust the noisy displacement for collision avoidance and traversability. We add the adjusted displacement to $x_t^i$ to obtain $x_{t+1}^i$. The velocity for the next step is computed from the updated positions as $v_{t+1}^i=(x_{t+1}^i-x_t^i)/\Delta t$.

\section{Crowd Generation}
\label{app:generation}

\subsection{Origin and Attribute Sampling}

To sample an origin, we apply the sigmoid function to $Z_{\mathrm{all}}$, mask the resulting scores with $M$, and normalize over grid cells:
\begin{equation}
p_o(u\mid S,M)
=\frac{\sigma(Z_{\mathrm{all}}(u))M(u)}
{\sum_{v\in\Omega}\sigma(Z_{\mathrm{all}}(v))M(v)}.
\label{eq:app_origin_sampling}
\end{equation}
For an arriving agent $i$, we sample a grid cell $u^i$ from $p_o$ and convert its location to the world-coordinate origin $o^i$.

Scene cross-attention at this origin predicts a categorical distribution for each of the four attribute groups. We sample one category from each distribution and concatenate the corresponding one-hot vectors to form $a^i$. We then compute the attribute embedding $e^i=W_a a^i$. Origins and attributes are thus sampled sequentially as in \cref{eq:emission}.

\subsection{Goal Sampling and Resampling}
\label{app:goal_sampling}

Goal sampling combines the goal head's logits with traversability penalties, a distance prior, and stationary occupancy scores on a $320\times320$ grid. We use the raw logits $Z_{\mathrm{goal}}^i$ underlying the predicted heatmap when constructing this distribution.

The stationary occupancy scores are
$P_{\mathrm{occ}}=\sigma(Z_{\mathrm{occ}})$.
They describe locations where pedestrians tend to stop and are computed from the emission features with zero position heatmaps. We normalize these scores by their maximum:
\begin{equation}
\widetilde{P}_{\mathrm{occ}}(u)
=
\frac{P_{\mathrm{occ}}(u)}
{\max_{v\in\Omega}P_{\mathrm{occ}}(v)}.
\label{eq:app_normalized_occupancy}
\end{equation}

To discourage nearby goals, we sample a distance threshold using origin-to-goal distance statistics from the training split. Let $\mathcal{D}_{\mathrm{goal}}$ denote these distances in grid-coordinate units, and let $Q_q$ denote the empirical quantile at level $q$. At each goal-sampling event, we select a quantile level $q^i\in\{0.25,0.5,0.75\}$ and draw a multiplier $\kappa^i\sim\mathcal{U}(0.8,1.2)$. The resulting distance threshold is
\begin{equation}
d_{\min}^i
=
\max\!\left\{
8,\,
0.6\,\kappa^i Q_{q^i}(\mathcal{D}_{\mathrm{goal}})
\right\}.
\label{eq:app_min_goal_distance}
\end{equation}
The distance threshold is at least eight pixels on the goal grid. Selecting different quantiles and multipliers varies this threshold across goal-sampling events.

We construct the adjusted sampling logits as
\begin{equation}
\begin{aligned}
\ell_g^i(u)
={}&Z_{\mathrm{goal}}^i(u)
+12\big(M(u)-1\big)\\
&+2\log\!\left(
\widetilde{P}_{\mathrm{occ}}(u)+10^{-3}
\right)\\
&-10^4\,\mathbf{1}\!\left[
\|u-\psi(x_t^i)\|_2<d_{\min}^i
\right].
\end{aligned}
\label{eq:app_goal_sampling_logits}
\end{equation}
Here, $u$ is interpreted as the two-dimensional coordinate of the candidate grid cell, and $\mathbf{1}[\cdot]$ is the indicator function.

Non-traversable cells receive a logit penalty of $12$. Cells closer than $d_{\min}^i$ receive a penalty of $10^4$, strongly suppressing their sampling probabilities relative to cells outside the threshold. The occupancy term favors locations where pedestrians tend to stop. The additive constant $10^{-3}$ keeps this term finite when the occupancy score is zero.

The goal distribution for agent $i$ is
\begin{equation}
p_g^i(u)
=
\frac{\exp\!\left(\ell_g^i(u)/\tau_g\right)}
{\sum_{v\in\Omega}\exp\!\left(\ell_g^i(v)/\tau_g\right)},
\qquad \tau_g=0.5.
\label{eq:app_goal_distribution}
\end{equation}
Here, $p_g^i$ denotes the grid-cell distribution for the current goal-sampling event. We sample one grid cell and convert its location to world coordinates.

At emission, $x_{t_0^i}^i=o^i$, and the sampled location becomes the initial goal $g^i$ in \cref{eq:emission}. When the agent reaches its current goal, we reuse the goal logits computed at emission, draw a new distance threshold, and evaluate the distance penalty relative to the agent's current position before sampling its next goal. The categorical attributes $a^i$ and embedding $e^i=W_a a^i$ remain unchanged throughout the agent's lifetime.

\subsection{Population Scheduling and Initialization}

We use statistics computed from the training split to sample scene population, group sizes, and agent lifetimes and to schedule arrivals. The arrival schedule determines when and how many agents enter the scene. Each sampled lifetime sets an upper bound on how long the corresponding agent remains active.

The learned emission module supplies each agent's origin, attributes, and initial goal. Each new agent is initialized at $o^i$ with initial goal $g^i$ and categorical attributes $a^i$. Its attribute embedding is computed as $e^i=W_a a^i$.

\subsection{Autoregressive Motion and Geometric Corrections}

At each step, the model predicts $\Delta_t^i$ for all active agents using their current scene features, states, and social context. We add temporally correlated noise to obtain the tentative next position:
\begin{equation}
\widetilde{x}_{t+1}^i
=x_t^i+\Delta_t^i+\epsilon_t^i.
\label{eq:app_rollout_update}
\end{equation}
The noise evolves as
\begin{equation}
\epsilon_t^i
=\rho\epsilon_{t-1}^i+\xi_t^i,
\label{eq:app_noise}
\end{equation}
where $\rho$ controls temporal dependence and $\xi_t^i$ is newly sampled noise. Both $\epsilon_t^i$ and $\xi_t^i$ have the same coordinate units as $\Delta_t^i$.

Let $\widetilde{\mathbf{x}}_{t+1}$ collect the tentative positions of all active agents. We first apply collision correction and then traversability correction:
\begin{equation}
\mathbf{x}_{t+1}
=\mathcal{W}_M\!\left(
\mathcal{C}_{\mathrm{col}}(\widetilde{\mathbf{x}}_{t+1})
\right).
\label{eq:app_correction_order}
\end{equation}
Here, $\mathcal{C}_{\mathrm{col}}$ jointly adjusts agent positions to address inter-agent collisions, and $\mathcal{W}_M$ adjusts positions according to the traversable-region mask $M$. Both corrections are applied after the learned motion prediction.

We compute the velocity for the next step from the corrected positions:
\begin{equation}
v_{t+1}^i
=\frac{x_{t+1}^i-x_t^i}{\Delta t}.
\label{eq:app_velocity_update}
\end{equation}
For the next prediction, we update the target and neighbor position heatmaps using the corrected positions and append the resulting velocities to the agents' velocity histories. We compute relative positions for social attention from the corrected positions in normalized coordinates.

Goals are resampled when reached, while $a^i$ and $e^i=W_a a^i$ remain fixed. Agents are removed when they leave the scene or their sampled lifetimes expire.

\section{Training Details and Losses}
\label{app:losses}

\subsection{Training Inputs and Gradient Flow}

We jointly train the emission and motion modules through the shared encoder. Emission predictions are supervised using annotated origins, attributes, goal locations, and stationary positions. The emission features are computed with zero position heatmaps, as in \cref{eq:app_emission_features}. Goal prediction applies FiLM to the global scene features using embeddings of the annotated attributes, as in \cref{eq:app_goal_film}.

During motion training, rollouts are initialized from annotated initial positions and attributes, together with the available motion histories. Annotated goal waypoints are provided at each step. The categorical attributes are mapped to embeddings using $e^i=W_a a^i$, as in \cref{eq:app_attribute_code}. Predicted positions are fed back to update velocities, position heatmaps, and social context autoregressively.

Because motion training uses annotated inputs, motion losses do not backpropagate through origins, attribute categories, or goals sampled by the emission module. They update the parameters used for motion prediction, including the shared encoder and attribute projection $W_a$. The origin, attribute-prediction, and goal heads receive their corresponding supervised emission losses. We use truncated backpropagation to limit gradient propagation through successive rollout states.

During generation, the motion module instead starts from origins, attributes, and goals sampled by the emission module.

Following \cref{eq:generator_objective}, the generator objective is
\begin{equation}
\mathcal{L}_G
=\mathcal{L}_{\mathrm{ce}}
+\mathcal{L}_{\mathrm{traj}}
+\mathcal{L}_{\mathrm{beh}}.
\label{eq:app_generator_objective}
\end{equation}
We sum these three loss groups without additional weighting. The component weights and averaging procedure within each group are specified below.

\Cref{alg:training} summarizes the training procedure, including supervised emission prediction, autoregressive motion rollouts, gradient truncation, and generator and discriminator updates.

\begin{algorithm}[t]
\caption{Training procedure of ACG}
\label{alg:training}
\footnotesize
\begin{algorithmic}[1]
\Require Annotated training data, generator parameters
$\Theta_G$, and discriminator parameters $\Theta_D$
\Require Rollout length $T_{\mathrm{roll}}$ and truncation
interval $K_{\mathrm{trunc}}$

\For{each training minibatch}
    \State Compute $e^i=W_a a^i$ from annotated attributes.

    \State $(F_{\mathrm{global}},F_{\mathrm{spatial}})
    \gets \theta_{\mathrm{enc}}([M,S,0,0])$

    \State Predict $Z_{\mathrm{all}}$,
    $Z_{\mathrm{attr}}$, and $Z_{\mathrm{occ}}$.

    \State Predict attribute distributions at annotated
    origins using scene cross-attention.

    \State Condition $F_{\mathrm{global}}$ on $e^i$
    using \cref{eq:app_goal_film} and predict goal heatmaps.

    \State Compute $\mathcal{L}_{\mathrm{ce}}$ using
    \cref{eq:app_emission_objective}.

    \State Initialize $\mathcal{S}_0$ from annotated
    initial positions and motion histories.

    \For{$t=0,\ldots,T_{\mathrm{roll}}-1$}
        \State Provide annotated goal waypoints $g_t^i$.

        \State Compute scene and social features from
        the current state $\mathcal{S}_t$.

        \State Form $z_t^i$ using
        \cref{eq:app_motion_input}.

        \State $\Delta_t^i
        \gets \exp(w^\top e^i)\,
        \phi_{\mathrm{mot}}(z_t^i,e^i)$

        \State Update $\mathcal{S}_{t+1}$ using
        the predicted displacements.

        \State Retain valid predicted positions and
        velocities for loss computation.

        \If{$(t+1)\bmod K_{\mathrm{trunc}}=0$}
            \State $\mathcal{S}_{t+1}
            \gets \operatorname{sg}(\mathcal{S}_{t+1})$
        \EndIf
    \EndFor

    \State Form the observed and predicted sequences of seven
    grid-normalized displacements from eight consecutive positions.
    
    \State Update $\Theta_D$ using \cref{eq:app_discriminator_objective}
    with detached displacement sequences and the categorical
    attributes $a^i$ concatenated at each step.
    
    \State Evaluate $\mathcal{L}_{\mathrm{adv}}$ on the predicted
    sequences using the updated discriminator.
    
    \State Compute $\mathcal{L}_{\mathrm{traj}}$ and
    $\mathcal{L}_{\mathrm{beh}}$ using
    \cref{eq:app_trajectory_objective,eq:app_behavior_components}.
    
    \State Update $\Theta_G$ using
    $\mathcal{L}_G=\mathcal{L}_{\mathrm{ce}}
    +\mathcal{L}_{\mathrm{traj}}+\mathcal{L}_{\mathrm{beh}}$.
\EndFor
\end{algorithmic}
\end{algorithm}

In \cref{alg:training}, $\mathcal{S}_t$ contains the rollout positions, velocities, motion histories, and position heatmaps. Observed trajectories provide supervision, while predicted positions update the rollout state. For the discriminator, $\widehat{\mathbf{v}}^{\,i}$ denotes the sequence of seven predicted displacements in grid coordinates, each divided by the grid width $G=320$.

The operator $\operatorname{sg}$ denotes stop-gradient. At each truncation boundary, we detach the state passed to subsequent steps. Stored predictions retain their computation graphs within each segment for loss computation.

During generator updates, discriminator parameters remain fixed while gradients propagate through its inputs to the generator. During discriminator updates, the generated sequences of grid-normalized displacements are detached, and only the discriminator parameters $\Theta_D$ are optimized.

\subsection{Emission Supervision}

\paragraph{Weighted heatmap supervision.}
For a predicted logit map $Z$ and target heatmap $Y$, let $P(u)=\sigma(Z(u))$. Both spawn and goal heatmap supervision use binary cross-entropy with a positive-class weight of $50$:
\begin{equation}
\begin{aligned}
\mathcal{B}_{50}(Z,Y)
&=-\frac{1}{|\Omega|}
\sum_{u\in\Omega}\Big[
50Y(u)\log P(u)\\
&\qquad +(1-Y(u))\log(1-P(u))
\Big].
\end{aligned}
\label{eq:app_map_bce}
\end{equation}
The loss is averaged over all grid cells. We use a $320\times320$ grid, giving $|\Omega|=320^2$.

\paragraph{Origin and stationary occupancy maps.}
Let $C_{\mathrm{orig}}$ denote the number of attribute-specific origin maps. The origin head predicts these maps together with one all-origin map and one stationary occupancy map.

The target $Y_{\mathrm{all}}$ is constructed from annotated starting locations. For each attribute-specific channel $c\in\{1,\ldots,C_{\mathrm{orig}}\}$, the target $Y_{\mathrm{attr},c}$ uses the starting locations of agents annotated with the corresponding category. Within each origin channel, Gaussian heatmaps centered at the annotated locations are combined using a pointwise maximum. Agents already present in the first frame are excluded from both the all-origin and attribute-specific origin targets.

The stationary occupancy target $Y_{\mathrm{occ}}$ uses positions at which the one-step displacement is below $0.15$\,m, sampled every two steps. Agents present in the first frame can also contribute to this target.

We average the spawn loss over the origin head's output channels.
\begin{equation}
\begin{aligned}
\mathcal{L}_{\mathrm{spawn}}
=\frac{1}{C_{\mathrm{orig}}+2}\Big[
&\mathcal{B}_{50}(Z_{\mathrm{all}},Y_{\mathrm{all}})\\
&+\sum_{c=1}^{C_{\mathrm{orig}}}
\mathcal{B}_{50}(Z_{\mathrm{attr},c},Y_{\mathrm{attr},c})\\
&+\mathcal{B}_{50}(Z_{\mathrm{occ}},Y_{\mathrm{occ}})
\Big].
\end{aligned}
\label{eq:app_origin_losses}
\end{equation}
The factor $1/(C_{\mathrm{orig}}+2)$ averages over output channels, in addition to the spatial averaging in \cref{eq:app_map_bce}. For minibatches containing multiple map tensors, we also average over those tensors. Stationary occupancy supervision is therefore included in $\mathcal{L}_{\mathrm{spawn}}$.

\paragraph{Attribute distributions.}
For predicted category probabilities $\pi_k^{i,(r)}$ and annotated one-hot labels $a^{i,(r)}$, the categorical cross-entropy is
\begin{equation}
\mathcal{L}_{\mathrm{attr}}
=-\mathbb{E}_i\!\left[
\sum_{r=1}^{4}\sum_{k=1}^{C_r}
a_k^{i,(r)}\log\pi_k^{i,(r)}
\right].
\label{eq:app_attribute_loss}
\end{equation}
The expectation is taken over supervised agents at their annotated origins. This loss directly supervises the predicted category distributions using the annotated labels, without requiring category sampling.

\paragraph{Goal heatmap supervision.}
Let $Z_{\mathrm{goal}}^i$ denote the goal logits obtained from attribute-conditioned emission features, as described in \cref{app:goal_prediction}. Let $y_{\mathrm{goal}}^i$ denote agent $i$'s annotated position at the end of the training window, converted to grid coordinates. We construct the target heatmap using a Gaussian with standard deviation $\sigma_g=4$ pixels:
\begin{equation}
Y_{\mathrm{goal}}^i(u)
=
\exp\!\left(
-\frac{\|u-y_{\mathrm{goal}}^i\|_2^2}
{2\sigma_g^2}
\right).
\label{eq:app_goal_target}
\end{equation}

For a minibatch of $N$ agents, the goal head is supervised by the weighted binary cross-entropy between its predicted heatmaps and the Gaussian targets:
\begin{equation}
\mathcal{L}_{\mathrm{goal}}
=\frac{1}{N}\sum_{i=1}^{N}
\mathcal{B}_{50}\!\left(Z_{\mathrm{goal}}^i,Y_{\mathrm{goal}}^i\right).
\label{eq:app_goal_loss}
\end{equation}
Here $\mathcal{B}_{50}$ averages over grid cells, and the outer mean averages over all agents in the minibatch without applying an alive mask.

During motion training, annotated goal waypoints are supplied as inputs to the motion module.

\paragraph{Combined emission objective.}
The emission objective combines binary cross-entropy for the origin, stationary occupancy, and goal maps with categorical cross-entropy for the four attribute groups.
\begin{equation}
\begin{aligned}
\mathcal{L}_{\mathrm{ce}}
={}&\lambda_{\mathrm{spawn}}\mathcal{L}_{\mathrm{spawn}}
+\lambda_{\mathrm{attr}}\mathcal{L}_{\mathrm{attr}}\\
&+\lambda_{\mathrm{goal}}\mathcal{L}_{\mathrm{goal}}.
\end{aligned}
\label{eq:app_emission_objective}
\end{equation}
We set $\lambda_{\mathrm{spawn}}=1.0$, $\lambda_{\mathrm{attr}}=0.5$, and $\lambda_{\mathrm{goal}}=1.0$. Stationary occupancy supervision is included in $\mathcal{L}_{\mathrm{spawn}}$.

\subsection{Trajectory Reconstruction}
Let $\mathcal{X}^i$ and $\widehat{\mathcal{X}}^i$ denote the observed and predicted trajectories over the supervised portion of a rollout. These trajectory sequences are distinct from the agent-token matrix $X_t$ used in social attention.

Let $\mathcal{P}$ denote the set of valid agent--time pairs. In the loss definitions, we write observed positions and velocities as $x_t^i,v_t^i$ and their predicted counterparts as $\widehat{x}_t^i,\widehat{v}_t^i$. The time-aligned position loss is
\begin{equation}
\mathcal{L}_{\mathrm{pos}}
=\frac{1}{|\mathcal{P}|}
\sum_{(i,t)\in\mathcal{P}}
\|\widehat{x}_t^i-x_t^i\|_1.
\label{eq:app_position_loss}
\end{equation}
Positions are measured in world coordinates, and the $\ell_1$ norm sums the absolute errors over the two coordinate dimensions.

We also use soft-DTW~\cite{cuturi2017soft} for differentiable temporal alignment. In the CAT configuration, each predicted and observed sequence contains eight absolute positions. Omitting the agent index, let $\widehat{u}_k$ and $u_l$ denote the predicted and observed positions in grid pixels. The local cost is $C_{kl}=\|\widehat{u}_k-u_l\|_2/G$, where $G=320$ is the grid width. Let $\mathcal{A}_{n,m}$ denote the set of binary matrices representing valid monotone alignment paths between the two sequences. The soft-DTW objective is
\begin{equation}
\begin{aligned}
&\operatorname{sDTW}_{\eta}
(\widehat{\mathcal{X}},\mathcal{X})\\
&\quad=-\eta\log
\sum_{A\in\mathcal{A}_{n,m}}
\exp\!\left(-\frac{\langle A,C\rangle}{\eta}\right),
\end{aligned}
\label{eq:app_soft_dtw}
\end{equation}
where $\eta>0$ controls smoothing and $\langle A,C\rangle=\sum_{k,l}A_{kl}C_{kl}$ is the total cost along an alignment path. We set $n=m=8$ and $\eta=1$ and use the soft-DTW value in \cref{eq:app_soft_dtw} without length normalization.

We average the soft-DTW objective over supervised trajectory pairs and combine it with the time-aligned position loss:
\begin{equation}
\begin{aligned}
\mathcal{L}_{\mathrm{sDTW}}
&=\mathbb{E}_i\!\left[
\operatorname{sDTW}_{\eta}
(\widehat{\mathcal{X}}^i,\mathcal{X}^i)
\right],\\
\mathcal{L}_{\mathrm{traj}}
&=\lambda_{\mathrm{pos}}\mathcal{L}_{\mathrm{pos}}
+\lambda_{\mathrm{sDTW}}\mathcal{L}_{\mathrm{sDTW}}.
\end{aligned}
\label{eq:app_trajectory_objective}
\end{equation}
We set $\lambda_{\mathrm{pos}}=1.0$ and $\lambda_{\mathrm{sDTW}}=0.5$.

\subsection{Behavior Distribution Matching}

\paragraph{Attribute-conditioned adversarial supervision.}
For each agent $i$, let $\mathbf{v}^i$ and
$\widehat{\mathbf{v}}^{\,i}$ denote the observed and predicted
sequences of seven grid-normalized displacements. For example,
the observed displacement at step $k$ is
$(u_{k+1}^i-u_k^i)/G$. At each step, the discriminator concatenates
the displacement with the 11-dimensional categorical attribute
vector $a^i$. A GRU with hidden size 128 encodes the resulting
sequence. The score head consists of a linear layer, a LeakyReLU
with negative slope 0.2, and a final linear layer.

Let $m_i$ be the binary alive indicator for agent $i$, and let
$w_i=m_i/\sum_j m_j$ be its weight among the agents in the batch.
The LSGAN~\cite{mao2017least} generator objective is
\begin{equation}
\mathcal{L}_{\mathrm{adv}}
=\sum_i w_i
\left(D(\widehat{\mathbf{v}}^{\,i},a^i)-1\right)^2.
\label{eq:app_generator_adversarial}
\end{equation}

The discriminator uses target scores 1 and 0 for observed and
generated sequences, respectively:
\begin{equation}
\begin{aligned}
\mathcal{L}_D
=\sum_i w_i\Big[&
\left(D(\mathbf{v}^i,a^i)-1\right)^2\\
&+D\!\left(\operatorname{sg}(\widehat{\mathbf{v}}^{\,i}),a^i\right)^2
\Big].
\end{aligned}
\label{eq:app_discriminator_objective}
\end{equation}
We first update the discriminator using detached sequences and
Adam with learning rate $2\times10^{-4}$ and
$(\beta_1,\beta_2)=(0.5,0.9)$. During the subsequent generator
update, gradients pass through the discriminator to the predicted
displacements. The discriminator is used only during training.

\paragraph{Attribute-wise speed matching.}
We use the quantile levels
$\mathcal{T}=\{0.1,0.25,0.5,0.75,0.9\}$ for both speed and spacing.
For two nonempty sample collections $U$ and $V$, define
\begin{equation}
\ell_Q(U,V)
=\frac{1}{5}\sum_{\tau\in\mathcal{T}}
\left|Q_\tau(U)-Q_\tau(V)\right|,
\label{eq:app_quantile_difference}
\end{equation}
where $Q_\tau$ denotes the empirical quantile at level $\tau$.

For each of the 11 attribute slots $c$, let
$\mathcal{I}_c=\{i:m_i=1,\ a_c^i=1\}$ be the alive agents
with that attribute. We retain the slots
$\mathcal{C}_3=\{c:|\mathcal{I}_c|\geq3\}$.
For each retained slot, we collect the observed step speeds
$\|u_{k+1}^i-u_k^i\|_2/G$ over all $i\in\mathcal{I}_c$ and
$k=1,\ldots,7$ in $\mathcal{V}_c$. We form
$\widehat{\mathcal{V}}_c$ analogously from predicted positions.
Repeated samples are retained. The speed loss averages the
quantile differences over the retained slots:
\begin{equation}
\mathcal{L}_{\mathrm{spd}}
=\frac{1}{|\mathcal{C}_3|}
\sum_{c\in\mathcal{C}_3}
\ell_Q(\widehat{\mathcal{V}}_c,\mathcal{V}_c),
\qquad |\mathcal{C}_3|>0.
\label{eq:app_speed_loss}
\end{equation}

\paragraph{Window-wise nearest-neighbor spacing.}
Let $\mathcal{I}_b$ be the alive agents in scene window $b$,
and let $\mathcal{B}_3=\{b:|\mathcal{I}_b|\geq3\}$ denote the
retained windows. For each agent $i\in\mathcal{I}_b$ and step
$k=1,\ldots,8$, we compute the observed and predicted distances
to the nearest other agent in the same window and step:
\begin{equation}
\begin{aligned}
d_{bik}
&=\frac{1}{G}\min_{j\in\mathcal{I}_b\setminus\{i\}}
\|u_k^i-u_k^j\|_2,\\
\widehat{d}_{bik}
&=\frac{1}{G}\min_{j\in\mathcal{I}_b\setminus\{i\}}
\|\widehat{u}_k^i-\widehat{u}_k^j\|_2.
\end{aligned}
\label{eq:app_neighbor_distances}
\end{equation}

Within each window, we collect the distances across agents and
steps in $\mathcal{R}_b$ and $\widehat{\mathcal{R}}_b$ for the
observed and predicted trajectories, respectively. We compute
the mean absolute quantile difference within each window and
then average over the retained windows:
\begin{equation}
\mathcal{L}_{\mathrm{nn}}
=\frac{1}{|\mathcal{B}_3|}
\sum_{b\in\mathcal{B}_3}
\ell_Q(\widehat{\mathcal{R}}_b,\mathcal{R}_b),
\qquad |\mathcal{B}_3|>0.
\label{eq:app_spacing_loss}
\end{equation}

\paragraph{Behavior loss weights.}
Following \cref{eq:beh}, we combine adversarial supervision, attribute-wise speed matching, and nearest-neighbor spacing as
\begin{equation}
\begin{aligned}
\mathcal{L}_{\mathrm{beh}}
={}&\lambda_{\mathrm{adv}}\mathcal{L}_{\mathrm{adv}}
+\lambda_{\mathrm{spd}}\mathcal{L}_{\mathrm{spd}}\\
&+\lambda_{\mathrm{nn}}\mathcal{L}_{\mathrm{nn}}.
\end{aligned}
\label{eq:app_behavior_components}
\end{equation}
For the CAT configuration, we use $\lambda_{\mathrm{adv}}=0.3$,
$\lambda_{\mathrm{spd}}=0.5$, and $\lambda_{\mathrm{nn}}=0.3$.

\section{Feature Space Comparisons}
\label{app:feature_comparison}

\paragraph{Evaluation setting.}
To assess motion patterns beyond the attribute-dependent motion statistics reported in the main paper, We compare real trajectories from CAT with trajectories generated by models trained on CAT, using the feature spaces of four pretrained trajectory predictors. Resonance~\cite{resonance} models social interactions through trajectory spectra, AgentFormer~\cite{agentformer} models temporal and social dependencies, GroupNet~\cite{groupnet} captures interactions at multiple scales, and Y-Net~\cite{ynet} incorporates scene context into goal and path prediction.

For both real and generated trajectories, each predictor receives eight input positions per agent. We extract features without providing pedestrian attribute labels to the predictors. We then use the labels to group the features into four pedestrian groups: adults on foot, children, personal mobility users, and pedestrians using phones. For the CrowdES baseline, we use a model retrained with pedestrian attribute annotations.

For each predictor and pedestrian group, we compute the squared maximum mean discrepancy ($\mathrm{MMD}^2$)~\cite{mmd} and Fr\'echet distance (FD)~\cite{frechet} between generated and real feature distributions. We compute each distance within an evaluation fold and report its mean over folds meeting the minimum sample requirement. This yields 16 comparisons per metric. We compare ACG and CrowdES within each feature space because the distance scales differ across predictors.

\paragraph{Results.}
As shown in \cref{tab:feat}, ACG achieves lower $\mathrm{MMD}^2$ than CrowdES in all 16 comparisons and lower FD in 15 of 16. For adults on foot, children, and personal mobility users, both metrics favor ACG in all four feature spaces. The largest relative reduction in $\mathrm{MMD}^2$ occurs for personal mobility users in the Y-Net feature space, from 0.234 for CrowdES to 0.090 for ACG (61.5\%). FD also decreases for this group, from 12.0 to 6.4.

For pedestrians using phones, $\mathrm{MMD}^2$ favors ACG in all four feature spaces, while FD favors ACG in three. In the Resonance feature space, FD decreases from 343.2 for CrowdES to 33.9 for ACG (90.1\%). Y-Net yields the only exception: CrowdES has the lower FD (6.3 versus 7.3), while ACG has the lower $\mathrm{MMD}^2$ (0.104 versus 0.131).

\input{tables/feature_comparison}

\input{tables/nav_attribute_supp.tex}

\section{Runtime Evaluation}
\label{app:runtime}
We measure the generation times reported in \cref{tab:latency} on the Zara1 test sequence (\texttt{crowds\_zara01}), which spans 360.6\,s of recorded motion. We use random seed 0 and a workstation equipped with a single NVIDIA RTX A6000 GPU (48\,GB).

We time the \texttt{generate()} call, excluding model loading and metric computation. Each active agent contributes one agent-step per generation step. Let $N_t$ denote the number of active agents at step $t$ and $T_{\mathrm{gen}}$ the elapsed generation time in milliseconds. We report $T_{\mathrm{gen}}/\sum_t N_t$ in milliseconds per agent-step.

\begin{figure*}[t]
    \centering
    \includegraphics[width=\textwidth]{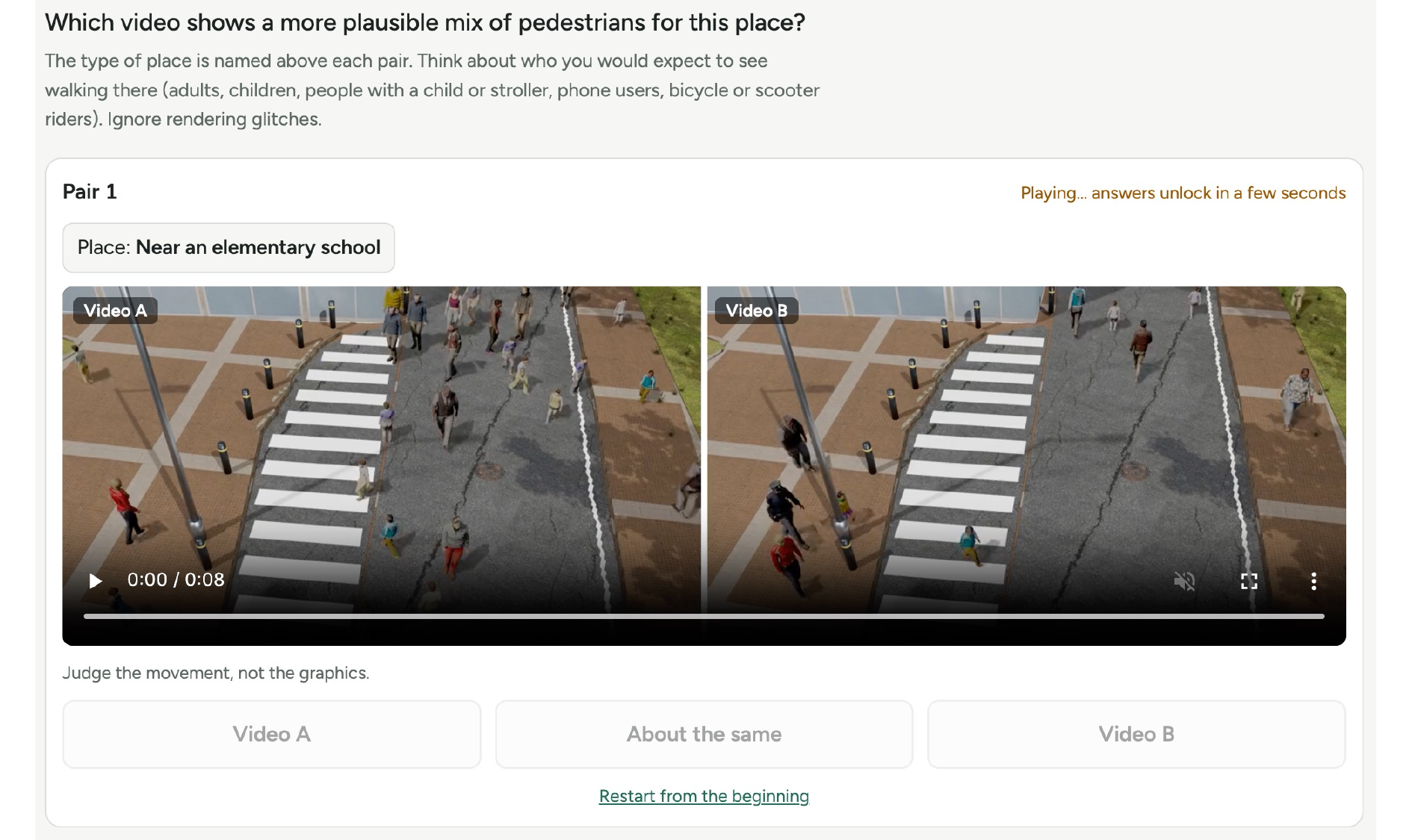}
    \caption{\textbf{User study interface.}}
    \label{fig:user_study_interface}
\end{figure*}

\section{Navigation Policy Evaluation by Pedestrian Attribute}
\label{app:nav_attribute}

In \cref{sec:nav}, we report the fraction of evaluation episodes in which the robot collides with a pedestrian. Here, we examine how often collisions occur in encounters with pedestrians from each attribute group. For a given group, the rate is the number of robot--pedestrian collisions involving members of that group per 100 encounters with members of the same group. \Cref{tab:nav_attr} compares this rate for policies trained on CrowdES and ACG crowds.

Policies trained on ACG crowds have lower collision rates in 15 of the 20 comparisons across four learned policies and five pedestrian groups. All four policies have fewer collisions per 100 encounters with adults and with phone users after training on ACG crowds. Three of the four policies also have lower rates for children, and three have lower rates for personal mobility users. For adults accompanying children, only RGL has a lower rate, decreasing from 1.04 to 0.56 collisions per 100 encounters.

\section{User Study}
\label{app:user_study}

\begin{figure*}[t]
    \centering
    \includegraphics[
        width=\textwidth,
        height=0.85\textheight,
        keepaspectratio
    ]{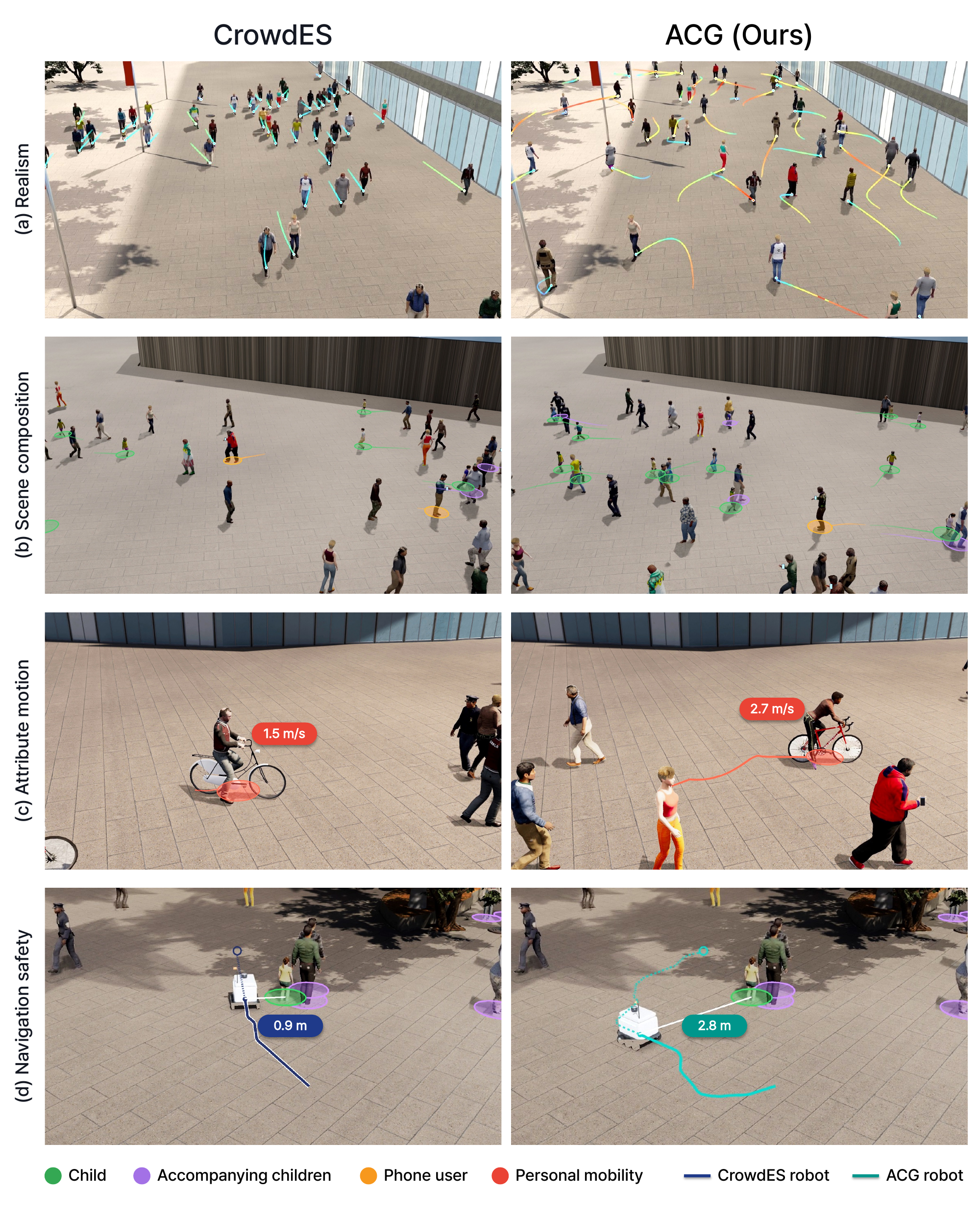}
    \caption{\textbf{Representative comparisons from the user study.} Each panel shows CrowdES on the left and ACG on the right. The four criteria are (a) crowd realism, (b) scene-dependent pedestrian composition, illustrated in a park scene, (c) attribute--motion consistency, illustrated with personal mobility users, and (d) perceived navigation safety of robot policies trained on crowds generated by each method. Colored ground rings indicate pedestrian attributes. The labels in (c) show pedestrian speeds, and those in (d) show robot--pedestrian distances.}
    \label{fig:user_study_example}
\end{figure*}

To evaluate the effectiveness of our method in generating diverse crowd simulation environments, we conduct a user study with 50 participants recruited through CloudResearch. We offered participants \$2.50 for a study advertised as taking 10--15 minutes. We did not seek ethics review because participants only watched simulated videos and selected responses, and we collected neither names nor contact information
Participants compare crowds generated by ACG and CrowdES and robot navigation policies trained on their respective generated crowds by answering four questions:
(1) Which video shows a more realistic crowd?
(2) Which video shows a more plausible mix of pedestrians for this place?
(3) In which video do the marked pedestrians move in a way that better fits who they are? 
(4) Which robot navigates around pedestrians more safely and appropriately?
These questions assess crowd realism, scene–attribute consistency, attribute--motion consistency, and perceived navigation safety, respectively.

We use three video pairs for each of the three crowd-generation criteria and nine pairs for navigation safety, collecting 900 responses across 18 video pairs.
Each pair is presented side by side as Video A and Video B, with randomized method placement to reduce positional bias.
Participants select Video A, About the same, or Video B.
For a fair comparison, paired videos use the same scene, camera viewpoint, and rendering procedure, with approximately comparable numbers of visible pedestrians. Location descriptions provide context for composition judgments, while colored ground rings identify pedestrians for attribute--motion evaluation.
The evaluation videos are rendered in CARLA~\cite{carla}, which is built on Unreal Engine~\cite{unrealsoftware}.
Participants are instructed to disregard rendering artifacts and focus on movement and behavior.

The questionnaire includes identical-video and instruction-following checks, requires at least 5\,s of viewing before a response, and records playback duration.
\cref{fig:user_study_interface} shows the participant instructions and an example video comparison, and \cref{fig:user_study_example} presents representative comparisons between CrowdES and ACG for each of the four evaluation criteria.
Preference percentages are computed separately for each criterion at the response level, with About the same counted as a tie.
The majority of responses favor ACG-generated crowds and navigation policies trained on them, supporting the effectiveness of our method for crowd simulation and navigation policy training.

\clearpage

%% file: tables/scene_statistics.tex
% Table: scene-level statistics (used inside a minipage; see main text)
\providecommand{\hd}[1]{\begin{tabular}[b]{@{}c@{}}#1\end{tabular}}
\centering
\caption{\textbf{Scene-level statistics.} Trajectory duration and speed are medians over tracks in each scene; speed is computed for tracks with at least 10 consecutive-frame steps. Observations are pedestrian positions sampled at 2 FPS.}
\label{tab:scene_statistics}
\setlength{\tabcolsep}{4pt}
\resizebox{\linewidth}{!}{%
\begin{tabular}{lrrrr}
\toprule
Scene & \# Pedestrians & \# Obs. & \hd{Traj.\\duration (s)} & \hd{Speed\\(m/s)} \\
\midrule
Univ. Canteen  &   192 &  11,420 & 23.00 & 1.38 \\
OCT Harbour    &   607 &  73,373 & 54.50 & 0.75 \\
Xili Crossroad &   503 &  50,228 & 49.00 & 1.07 \\
Primary School &   405 &  36,769 & 50.00 & 0.43 \\
Univ. Campus   &   346 &  36,463 & 54.00 & 0.92 \\
Huaqiangbei    & 1,995 & 170,147 & 27.00 & 1.26 \\
\midrule
Total / Pooled & 4,048 & 378,400 & 37.00 & 1.05 \\
\bottomrule
\end{tabular}%
}

%% file: tables/scene_attr_statistics.tex
\begin{table}[!t]
\centering
\caption{\textbf{Scene-wise attribute statistics of CAT.}  Share is the percentage of pedestrians in each scene. ``--'' denotes fewer than 10 pedestrians.}
\label{tab:scene_attr_stats}
\providecommand{\hd}[1]{\begin{tabular}[b]{@{}c@{}}#1\end{tabular}}
\providecommand{\sd}[1]{{\scriptsize$\pm$#1}}
\providecommand{\iq}[2]{{\scriptsize[#1, #2]}}
\footnotesize
\renewcommand{\arraystretch}{0.95}
\setlength{\tabcolsep}{2pt}
\resizebox{\columnwidth}{!}{%
\begin{tabular}{@{}lccccc@{}}
\toprule
 & \hd{Adult /\\on foot} & Child & \hd{Personal\\mobility} & \hd{Accomp.\\a child} & \hd{Phone\\use} \\
\midrule
\textit{Univ. Canteen} \\
\quad Share (\%) & 72.4 & 0.0 & 6.2 & 0.0 & 15.1 \\
\quad Speed (m/s) & 1.32\,\sd{0.31} & -- & -- & -- & 1.36\,\sd{0.49} \\
\quad Acceleration (m/s$^2$) & 0.15\,\sd{0.07} & -- & -- & -- & 0.17\,\sd{0.10} \\
\quad Stop time (s) & 5.5\,\iq{3.4}{8.3} & -- & -- & -- & 5.5\,\iq{3.8}{13.2} \\
\quad Neighbor dist.\ (m) & 0.95\,\iq{0.73}{1.83} & -- & -- & -- & 2.61\,\iq{1.19}{4.55} \\
\quad Companion (\%) & 54 & -- & -- & -- & 26 \\
\addlinespace[2pt]
\textit{OCT Harbour} \\
\quad Share (\%) & 51.7 & 16.5 & 0.3 & 17.8 & 5.6 \\
\quad Speed (m/s) & 0.85\,\sd{0.27} & 0.86\,\sd{0.28} & -- & 0.75\,\sd{0.22} & 0.80\,\sd{0.27} \\
\quad Acceleration (m/s$^2$) & 0.20\,\sd{0.12} & 0.19\,\sd{0.10} & -- & 0.18\,\sd{0.10} & 0.13\,\sd{0.06} \\
\quad Stop time (s) & 3.5\,\iq{2.5}{8.4} & 4.0\,\iq{3.0}{6.0} & -- & 4.3\,\iq{3.0}{7.9} & 8.0\,\iq{2.5}{14.2} \\
\quad Neighbor dist.\ (m) & 0.76\,\iq{0.57}{1.19} & 0.83\,\iq{0.57}{1.23} & -- & 0.84\,\iq{0.61}{1.23} & 1.21\,\iq{0.59}{2.21} \\
\quad Companion (\%) & 57 & 52 & -- & 58 & 47 \\
\addlinespace[2pt]
\textit{Xili Crossroad} \\
\quad Share (\%) & 57.3 & 10.1 & 14.1 & 8.5 & 3.6 \\
\quad Speed (m/s) & 1.13\,\sd{0.22} & 1.00\,\sd{0.23} & 2.19\,\sd{1.00} & 0.93\,\sd{0.18} & 1.11\,\sd{0.18} \\
\quad Acceleration (m/s$^2$) & 0.33\,\sd{0.16} & 0.39\,\sd{0.21} & 0.71\,\sd{0.36} & 0.33\,\sd{0.16} & 0.30\,\sd{0.13} \\
\quad Stop time (s) & 4.0\,\iq{2.5}{13.3} & 3.0\,\iq{2.5}{5.8} & 4.0\,\iq{2.5}{10.4} & 4.0\,\iq{2.5}{11.1} & 11.0\,\iq{4.1}{19.6} \\
\quad Neighbor dist.\ (m) & 0.81\,\iq{0.55}{1.42} & 0.85\,\iq{0.47}{1.54} & 1.82\,\iq{1.10}{2.62} & 0.68\,\iq{0.52}{0.89} & 1.10\,\iq{0.80}{1.56} \\
\quad Companion (\%) & 47 & 47 & 10 & 58 & 25 \\
\addlinespace[2pt]
\textit{Primary School} \\
\quad Share (\%) & 17.0 & 55.6 & 0.2 & 19.5 & 3.2 \\
\quad Speed (m/s) & 0.91\,\sd{0.38} & 0.89\,\sd{0.36} & -- & 0.95\,\sd{0.32} & -- \\
\quad Acceleration (m/s$^2$) & 0.22\,\sd{0.09} & 0.20\,\sd{0.08} & -- & 0.16\,\sd{0.05} & -- \\
\quad Stop time (s) & 36.8\,\iq{9.5}{73.5} & 10.0\,\iq{4.8}{28.0} & -- & 9.5\,\iq{6.0}{22.9} & 51.4\,\iq{15.3}{73.5} \\
\quad Neighbor dist.\ (m) & 0.73\,\iq{0.58}{1.00} & 0.43\,\iq{0.38}{0.58} & -- & 0.44\,\iq{0.40}{0.56} & 0.96\,\iq{0.80}{1.21} \\
\quad Companion (\%) & 41 & 79 & -- & 87 & -- \\
\addlinespace[2pt]
\textit{Univ. Campus} \\
\quad Share (\%) & 51.4 & 14.7 & 13.9 & 7.2 & 6.6 \\
\quad Speed (m/s) & 0.98\,\sd{0.27} & 0.90\,\sd{0.17} & 3.60\,\sd{1.31} & 0.91\,\sd{0.26} & 1.12\,\sd{0.29} \\
\quad Acceleration (m/s$^2$) & 0.16\,\sd{0.07} & 0.15\,\sd{0.10} & 0.32\,\sd{0.15} & 0.15\,\sd{0.06} & 0.12\,\sd{0.07} \\
\quad Stop time (s) & 5.5\,\iq{3.5}{12.1} & 6.5\,\iq{3.5}{8.0} & -- & 7.3\,\iq{3.9}{15.5} & 18.9\,\iq{7.6}{37.3} \\
\quad Neighbor dist.\ (m) & 0.87\,\iq{0.62}{1.22} & 0.61\,\iq{0.51}{0.76} & 3.49\,\iq{1.24}{4.54} & 0.85\,\iq{0.54}{1.08} & 2.36\,\iq{1.35}{3.92} \\
\quad Companion (\%) & 63 & 84 & 22 & 68 & 24 \\
\addlinespace[2pt]
\textit{Huaqiangbei} \\
\quad Share (\%) & 74.6 & 5.2 & 1.7 & 2.4 & 14.3 \\
\quad Speed (m/s) & 1.14\,\sd{0.35} & 0.91\,\sd{0.28} & 3.21\,\sd{0.85} & 0.91\,\sd{0.24} & 1.05\,\sd{0.41} \\
\quad Acceleration (m/s$^2$) & 0.44\,\sd{0.29} & 0.32\,\sd{0.22} & 0.66\,\sd{0.42} & 0.33\,\sd{0.20} & 0.36\,\sd{0.23} \\
\quad Stop time (s) & 2.5\,\iq{2.0}{5.8} & 2.9\,\iq{2.0}{6.4} & -- & 3.5\,\iq{2.0}{8.1} & 6.3\,\iq{2.8}{26.5} \\
\quad Neighbor dist.\ (m) & 1.52\,\iq{0.92}{2.05} & 0.94\,\iq{0.62}{1.54} & 1.86\,\iq{1.55}{2.05} & 1.12\,\iq{0.68}{1.81} & 1.79\,\iq{1.24}{2.18} \\
\quad Companion (\%) & 23 & 41 & 5 & 39 & 17 \\
\bottomrule
\end{tabular}%
}
\end{table}

%% file: tables/feature_comparison.tex
\begin{table}[t]
\centering
\footnotesize
\renewcommand{\arraystretch}{1.12}
\setlength{\tabcolsep}{2pt}
\begin{tabular*}{\columnwidth}{@{\extracolsep{\fill}}lcccc@{}}
\toprule
& \multicolumn{2}{c}{MMD$^2\downarrow$}
& \multicolumn{2}{c}{FD$\downarrow$} \\
\cmidrule(lr){2-3}\cmidrule(lr){4-5}
Pedestrian group & ACG & CrowdES & ACG & CrowdES \\
\midrule

\multicolumn{5}{l}{\textit{Resonance}~\cite{resonance}} \\
Adult / on foot
& \textbf{0.072} & 0.152 & \textbf{17.1} & 82.9 \\
Child
& \textbf{0.071} & 0.179 & \textbf{17.4} & 77.0 \\
Personal mobility
& \textbf{0.057} & 0.082 & \textbf{16.6} & 25.1 \\
Phone use
& \textbf{0.110} & 0.174 & \textbf{33.9} & 343.2 \\

\midrule

\multicolumn{5}{l}{\textit{AgentFormer}~\cite{agentformer}} \\
Adult / on foot
& \textbf{0.114} & 0.200 & \textbf{30.9} & 69.1 \\
Child
& \textbf{0.225} & 0.301 & \textbf{68.9} & 107.0 \\
Personal mobility
& \textbf{0.224} & 0.313 & \textbf{65.3} & 103.8 \\
Phone use
& \textbf{0.271} & 0.380 & \textbf{83.1} & 197.6 \\

\midrule

\multicolumn{5}{l}{\textit{GroupNet}~\cite{groupnet}} \\
Adult / on foot
& \textbf{0.122} & 0.215 & \textbf{17.5} & 22.3 \\
Child
& \textbf{0.199} & 0.280 & \textbf{11.9} & 31.6 \\
Personal mobility
& \textbf{0.262} & 0.415 & \textbf{22.8} & 51.7 \\
Phone use
& \textbf{0.185} & 0.274 & \textbf{18.9} & 30.5 \\

\midrule

\multicolumn{5}{l}{\textit{Y-Net}~\cite{ynet}} \\
Adult / on foot
& \textbf{0.075} & 0.134 & \textbf{5.1} & 15.1 \\
Child
& \textbf{0.082} & 0.147 & \textbf{4.6} & 7.5 \\
Personal mobility
& \textbf{0.090} & 0.234 & \textbf{6.4} & 12.0 \\
Phone use
& \textbf{0.104} & 0.131 & 7.3 & \textbf{6.3} \\

\bottomrule
\end{tabular*}
\caption{
\textbf{Feature space comparisons on CAT.}
MMD$^2$ and Fr\'echet distance (FD) measure differences
between generated and real feature distributions for
each predictor and pedestrian group.
CrowdES is retrained with pedestrian attribute annotations.
Lower values indicate closer agreement with real data,
and bold marks the lower value in each comparison.
}
\label{tab:feat}
\end{table}

%% file: tables/nav_attribute_supp.tex
\begin{table}[t]
\centering
\resizebox{\columnwidth}{!}{%
\begin{tabular}{llccccc}
\toprule
& & \multicolumn{5}{c}{Collisions per 100 encounters$\downarrow$} \\
\cmidrule(lr){3-7}
Policy & Training crowd & Adult & Child & PM & Accomp & Phone \\
\midrule

ORCA & -- & 6.99 & 3.07 & 15.44 & 3.46 & 5.88 \\

\midrule

SARL & CrowdES
& 1.75 & \textbf{0.60} & 8.75 & \textbf{0.56} & 1.72 \\
\rowcolor{oursgray}
& ACG
& \textbf{1.60} & 0.79 & \textbf{8.33} & 0.73 & \textbf{0.90} \\

\midrule

RGL & CrowdES
& 1.99 & 0.67 & \textbf{7.13} & 1.04 & 1.77 \\
\rowcolor{oursgray}
& ACG
& \textbf{1.79} & \textbf{0.55} & 9.69 & \textbf{0.56} & \textbf{1.55} \\

\midrule

LSTM-RL & CrowdES
& 1.67 & 0.83 & 8.95 & \textbf{0.47} & 1.58 \\
\rowcolor{oursgray}
& ACG
& \textbf{1.45} & \textbf{0.67} & \textbf{6.84} & 0.52 & \textbf{1.11} \\

\midrule

CrowdNav++ & CrowdES
& 3.66 & 1.54 & 11.66 & \textbf{1.43} & 3.19 \\
\rowcolor{oursgray}
& ACG
& \textbf{2.73} & \textbf{0.90} & \textbf{7.64} & 1.47 & \textbf{1.83} \\

\bottomrule
\end{tabular}}
\caption{
\textbf{Collisions by pedestrian attribute.}
Collisions per 100 encounters for policies trained on
CrowdES or ACG crowds, with ORCA included as a reference.
PM denotes pedestrians using mobility, and
Accomp denotes adults accompanying children.
Bold indicates the lower collision rate between the
two training crowds for each policy and pedestrian group.
}
\label{tab:nav_attr}
\end{table}